# Coalitions of AI-based Methods Predict 15-Year Risks of Breast Cancer Metastasis Using Real-World Clinical Data with AUC up to 0.9

Xia Jiang[1], Yijun Zhou[1], Alan Wells[2,4], Adam Brufsky[3,4]

[1]Department of Biomedical Informatics
University of Pittsburgh
Pittsburgh, PA

[2]Department of Pathology
University of Pittsburgh and Pittsburgh VA Health System
Pittsburgh, PA

[3]Division of Hematology/Oncology
University of Pittsburgh School of Medicine
Pittsburgh, PA

[4]UPMC Hillman Cancer Center
Pittsburgh, PA

**Contact:**  Xia Jiang

**Email:**  xij6@pitt.edu

**Phone:**  412-648-9310

Yijun Zhou: yiz209@pitt.edu
Alan Wells: wellsa@upmc.edu
Adam Brufsky: brufskyam@upmc.edu

## ABSTRACT

**Background**
Breast cancer is one of the two cancers responsible for the most deaths in women, with about 42,000 deaths each year in the US. That there are over 300,000 breast cancers newly diagnosed each year suggests that only a fraction of the cancers result in mortality. Thus, most of the women undergo seemingly curative treatment for localized cancers, but a significant later succumb to metastatic disease for which current treatments are only temporizing for the vast majority. The current prognostic metrics are of little actionable value for 4 of the 5 women seemingly cured after local treatment, and many women are exposed to morbid and even mortal adjuvant therapies unnecessarily, with these adjuvant therapies reducing metastatic recurrence by only a third. Thus, there is a need for better prognostics to target aggressive treatment at those who are likely to relapse and spare those who were actually 'cured'. While there is a plethora of molecular and tumor-marker assays in use and under-development to detect recurrence early, these are time consuming, expensive and still often un-validated as to actionable prognostic utility. A different approach would use large data techniques to determine clinical and histopathological parameters that would provide accurate prognostics using existing data. Herein, we report on machine learning, together with grid search and Bayesian Networks to develop algorithms that present a AUC of up to 0.9 in ROC analyses, using only extant data. Such algorithms could be rapidly translated to clinical management as they do not require testing beyond routine tumor evaluations.

**Methods**
By using open-source libraries provided through Scikit-Learn, we developed a python package named as RSGP for conducting grid search with ten *machine learning* (*ML*) methods including deep learning to carry on our specific prediction tasks. We previously developed MBIL, a *Bayesian network* (*BN*)-based method for identifying risks factors (RFs) for a disease outcome such as *breast cancer metastasis* (BCM). Using RSGP and MBIL, we developed two types of prediction models, the RF type, resulted from the coalition of a ML method, grid search, and MBIL, and the all-feature type, resulted from the coalition of a ML method and grid search. This was done for each of the ten ML methods, resulting in 20 best models, 10 for each type. We obtained such a set of 20 best models for predicting the risks of 5-year, 10-year, and 15-yeart *BCM* respectively, resulting in 60 best prediction models, and each model is selected from its grid searches that trained and tested 30000 different models following a 5-fold cross validation procedure. Mean-test AUC scores were obtained from grid searches and ROC_curves were developed based on independent validation tests.

**Results**
The DNM_RF model presents a mean-test AUC of 0.862, the highest mean-test AUC we have achieved in predicting future risks of BCM. Based on mean-test AUCs, the RF type of models outperforms the all-feature type for eight out of the ten ML methods when predicting the risks of 15-year BCM. The LASSO-15Year model yields a validation AUC of 0.901, the best AUC we have ever achieved for predicting future risks of BCM. The SHAP analyses reveal important predictors for 15-year BCM.

**Conclusion**
Grid search greatly improves prediction not only for deep learning but also for some other ML methods such as XGB and *Random Forrests*. The study demonstrates the effectiveness of the BN-based MBIL method for identifying risk factors for a disease outcome such as 15-year BCM. The coalitions among grid search, MBIL, and a ML method such as deep learning can be very powerful to improve prediction.

# INTRODUCTION

## About deep learning and grid search

Deep learning has become an important AI-based prediction method during the last two decades [1–4]. It is a machine learning model architecture that was developed based on the *Artificial Neural Network* (*ANN*). The ANN was originally developed to recognize patterns and conduct prediction using a model structure that consists of an input layer, an output layer, and a single hidden layer of nodes, which, in a loose manner, have similar functions such as receiving and sending out signals as the neurons in human nervous system [5,6]. Deep learning refers to a machine learning model architecture that stems out of the original ANN but consists of more than one hidden layers of nodes [1–4,7,8]. Deep learning has obtained significant success in commercialized applications such as voice and pattern recognition, computer vision, and image processing [9–19].

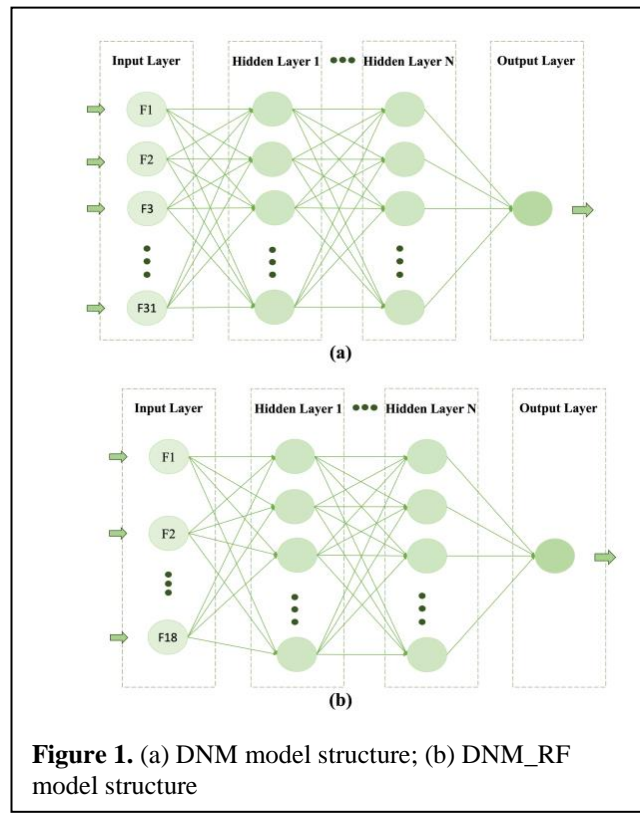

**Figure 1.** (a) DNM model structure; (b) DNM_RF model structure

**The DNM models:** The *Deep feed forward neural network (DFNN) models for predicting the future risk of breast cancer metastasis* (*BCM*) can be learned from non-image clinical data concerning breast cancer [20–23]. Although the DFNN method can target at other disease outcomes such as 5-year survival, we focus on BCM in this study. Therefore, we call our models *DNM* (*DFNN-BCM*). The datasets we used are two-dimensional, because they contain both columns and rows as we see in a common two-dimensional table. In such a dataset, a column often represents an attribute or a property, for example the stage of breast cancer, which is commonly called a variable or feature in the world of machine learning. A column contains the values of a feature from all the subjects. A row often represents a subject, for example a patient, which is commonly called a case or data point in the world of machine learning. A row contains the values of all the features for a particular subject. We will describe the specific datasets we use in this study in the Methods section below.

A DFNN-based model can to some extent be viewed as a "general case" of the traditional ANN model. Just like a traditional ANN model, a DFNN model contains one input layer and one output layer. But unlike a traditional ANN model that consists only one hidden layer, a DFNN model can contain one or more than one hidden layer(s). Figure 1 (a) shows the structure of our DFNN models. In these models, the input layer contains 31 nodes, representing the 31 clinical features other than the outcome feature, contained in our datasets. The output layer contains one node, which represents our binary outcome feature called *metastasis*. *Metastasis* has two values:

0 and 1. When it is equal to 0, no metastasis is found in patient, when it is equal to 1, metastasis is found in patient.

**Grid search:** The prediction performance of a DFNN-based model that is learned from data is closely associated with a learning scheme called grid search [22,23]. There is a large number of adjustable hyperparameters in a deep learning method like the DFNN, and different value assignments for the set of adjustable hyperparameters can result in models that perform differently. This can be considered as an advantage of deep learning because more adjustable hyperparameters allow us to have more ways of changing and improving a model. But on the other hand, having a large number of adjustable hyperparameters makes it a more challenging task to conduct hyperparameter tunning, which is the process of determining an optimal value assignment for all hyperparameters. A grid search can be considered as a systematic way of conducting hyperparameter tunning [22,24,25]. We describe the procedure of our grid searches as follows: firstly, we determine a set of values for each of the adjustable hyperparameters. For example, an adjustable hyperparameter called *learning rate* can technically take an infinite number of different values ranging from 0 to 1, therefore, we need to select a fixed number of values for *learning rate*; secondly, we give the preselected sets of values for the adjustable hyperparameters to our grid search program as one of its input; thirdly, we run our grid search program, which conducts an independent model training and testing process at every unique value assignment of the set of hyperparameters, determined by the sets of input hyperparameter values. Such an unique value assignment of the set of adjustable hyperparameters is called a *hyperparameter setting* (*HYPES*) in our research [22,23]; finally, our grid search program automatically stores as one of its put the HYPES and corresponding model performance scores resulted from each of the independent model training and testing processes.

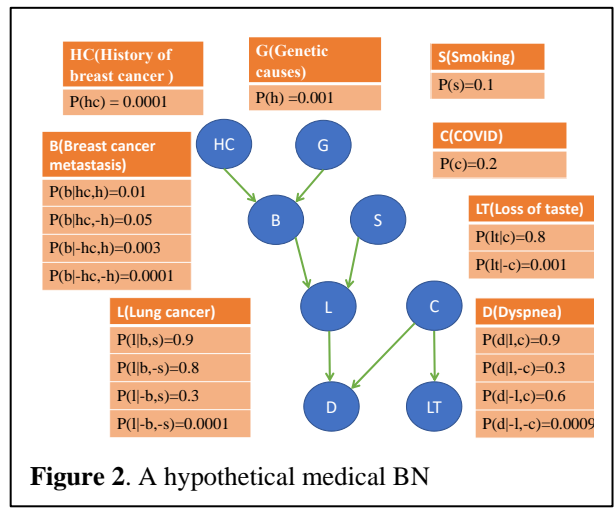

**Figure 2**. A hypothetical medical BN

**About Bayesian Networks**
**Bayesian networks:** *Bayesian networks* (*BNs*) have become a leading architecture for modeling uncertain reasoning in artificial intelligence and machine learning [26–30]. A Bayesian network consists of a directed acyclic graph (DAG), whose nodes represent random variables, and the conditional probability distribution of every variable in the network given each set of values of its parents [30]. We will use the terms node and variable interchangeably in this context. The directed edges represent direct probabilistic dependencies. In general, for each node $X_i$ there is a probability distribution on that node given the state of its parents, which are the nodes with edges going into $X_i$. The nodes that can be reached by following a directed path from $X_i$ (following a tail-to-head direction of the edges) are called the descendants of $X_i$. For example, in an 8-nodes hypothetical medical BN shown in Figure 2, node *D* has two parents, which are *L* and *C*, and node *S* has two descendants, which are *L* and *D*. A BN encodes a joint probability distribution, and therefore it represents all the information needed to compute any marginal or conditional probability on the nodes in the network. A variety of algorithms have been developed for rapidly computing P(XS1 | XS2), where XS1 and XS2 are arbitrary sets of variables with instantiated values [27,30,31].

**Markov Blanket and MBIL:** In a BN, a *Markov blanket* of a given node $T$ contains at least the set of nodes $M$ such that $T$ is probabilistically independent of all other nodes in the network conditional on the nodes in $M$ [30]. In general, a Markov blanket of $T$ contains at least all parents of $T$, children of $T$, and parents of children of $T$. If $T$ is a leaf (a node with no children), then a Markov blanket consists only of the parents of $T$. Figure 3 shows a BN DAG model. Since $T$ is a leaf in that BN, a Markov blanket $M$ of $T$ consists of its parents, namely nodes $X_{11}$-$X_{15}$. Without knowing the BN DAG model, nodes $X_1$-$X_{10}$ $X_{16}$, and $X_{17}$ would all be learned as predictors because they are indirectly connected to $T$ through the nodes in the Markov blanket $M$ [21]. However, if we can identify $M$ and know the values of the nodes in it, we will have blocked the connections between $T$ and the other nodes. So, the other nodes can be completely removed from our prediction model without affecting the prediction performance of the model. This helps reduce the complexity of a prediction model, and therefore, could hypothetically improve prediction performance and reduce computational cost, which is one of the challenges for deep learning with grid search [22,24,25,32]. Based on this idea, we previously developed the *Markov Blanket-based Interactive and Direct Risk Factor Learner* (*MBIL*) [21], a supervised BN-based method for learning causal risk factors for a target feature such as BCM. We developed MBIL based on the Bayesian network learning techniques and its relevant concepts such as Markov Blankets [21,30], as described above. MBIL can be used for identifying risk factors for a target feature [21]. It not only detects single causative risk factors in a Markov blanket of a leaf target node, but also interactive factors that jointly affect such a target node [21].

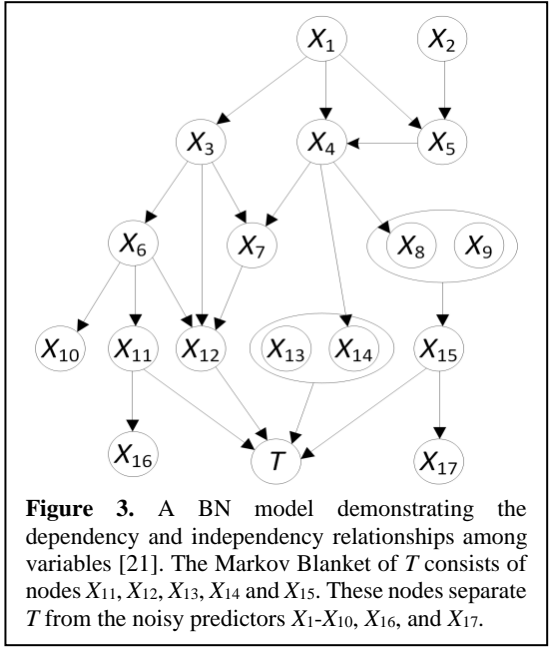

**Figure 3.** A BN model demonstrating the dependency and independency relationships among variables [21]. The Markov Blanket of $T$ consists of nodes $X_{11}$, $X_{12}$, $X_{13}$, $X_{14}$ and $X_{15}$. These nodes separate $T$ from the noisy predictors $X_1$-$X_{10}$, $X_{16}$, and $X_{17}$.

**About the purpose of this study**
In this study, we develop and optimize both DNM and *DNM_RF* models through well-designed grid searches. RF stands for risk factor. We develop DNM_RF models by applying the MBIL package that can be used to learn interactive and direct risk factors [21]. We first learn RFs that are predictive of BCM from our datasets that contain 31 clinical features ( see details in Table S1 of the supplement), then we retrieve new datasets using the RFs. The DNM_RF models are learned from the new datasets. Therefore, the input layer of the DNM_RF models contains less number of features than that of the DNM models, as demonstrated in Figure 1. We describe in details how the DNM and DNM_RF models are developed in the Methods section below.

The main purpose of this study is to compare the DNM models and the DNM_RF models. We also want to compare the DFNN-based models with some other typical machines learning (*ML*) models, which are all developed through grid searches. The set of ML methods we use in this study are described in Methods. We made the following assumptions: 1) *A DNM_RF prediction model will yield better predictive performance in grid search than the corresponding DNM prediction model;* and 2) *A DFNN-based prediction model will yield better predictive performance than the representative set of other ML-based prediction models.* We made these assumptions because 1) using the RFs that are found by the BN-based MBIL to guide prediction will help reduce model complexity and tease out the "noises" made by the non BCM-predictive features, and this could lead to a better prediction performance and reduce time for grid search; 2) DFNN has a large set of adjustable hyperparameters which allow sophisticated hyperparameter tuning to improve prediction and reduce overfitting; 3) Deep learning is a popular and powerful prediction tool as demonstrated by its successes in many other applications.

**METHODS**

**Datasets**
We use six datasets concerning BCM in this study. Among them, the *LSM-5Year*, *LSM-10Year*, and *LSM-15Year* were developed and made available through previous studies [20,21]. The *LSM_RF-5Year*, *LSM_RF-10Year*, and *LSM_RF-15Year* are developed using the RFs identified by the MBIL method [21]. Using the LSM_RF-5Year as an example, the 2-step procedure for curating this dataset is as follows: Step 1: Applying the MBIL method to the LSM-5Year to retrieve the RFs that are predictive of BCM. The LSM-5Year contains 32 variables including an outcome variable called *metastasis*, which represents the state of having or not having BCM by the 5$^{th}$ year post the initial treatment. Sometimes we call the outcome variable the target feature. The remaining 31 variables are the predictive features, which are also referred to as predictors when they are used to predict a patient outcome; Step 2: Removing from the LSM-5Year all columns of the non-outcome features that don't belong to the set of RFs found in Step 1, and the remaining part of the data forms the LSM_RF-5Year. We followed the same 2-step procedure to obtain the LSM_RF-10Year from the LSM-10Year and the LSM_RF-15 year from the LSM-15Year. Table 1 below shows the counts of the cases and predictors included in the six datasets. More detailed descriptions of the predictors for the six datasets are included in Table S1-S4 of the supplement.

**Table 1.** Case counts and number of predictors of the six LSM datasets

|   | Total # of cases | # of Positive cases | # of Negative cases | # of Predictors |
|---|---|---|---|---|
| **LSM-5year** | 4189 | 437 | 3752 | 31 |
| **LSM-10year** | 1827 | 572 | 1255 | 31 |
| **LSM-15year** | 751 | 608 | 143 | 31 |
| **LSM_RF-5year** | 4189 | 437 | 3752 | 20 |
| **LSM_RF-10Year** | 1827 | 572 | 1255 | 18 |
| **LSM_RF-15Year** | 751 | 608 | 143 | 17 |

**The RGS strategy, and the RGSP package for developing DNM and DNM_RF models**
We developed *the Randomized Grade Search Package (RGSP)* python package which can be used to develop and optimize the DFNN types of models. RGSP contains the following major components other than the routine dataset processor and output generator: 1) The DFNN model builder, which uses the libraries provided by the Keras package [33]. Keras is a high-level neural network API built on top of TensorFlow [33,34]. The Keras package is made available in a collection of python packages named as Scikit-Learn [33]. TensorFlow is an open-source development platform for machine learning and artificial intelligence (AI), and Keras can be viewed as a wrapper of TensorFlow. Such a wrapper serves as a communication interface between a deep learning developer and TensorFlow; 2) The DFNN model learner, which follows the k-fold *cross validation* (*CV*) strategy to train and test DFNN models by calling the DFNN model builder. Since the k-fold CV is also closely related to the evaluation of the DFNN models, more detailed information about this component is seen in the "Performance Metrics" subsection below; 3) The *Randomized Grid Search* (*RGS*) *Hyperparameter Setting Generator* (*RGS_HSG*), which takes as input a preselected set of values for each of the adjustable hyperparameters to produce randomly selected HYPESs. We call all possible HYPESs that are determined by the input sets of hyperparameter values the *pool of hyperparameter settings* (*P-HYPESs*). The number of HYPESs in the pool can be very large. Using the sets of hyperparameter values we used in our experiments concerning DFNN method (as shown in Table S5 of the supplement) as an example, the number of available unique HYPESs in the correspond P-HYPESs is the product of 332, 4189, 299, 90, 89, 299, 299, 299, 4, 4189, and 400. So running a "full" grid search, that is, using every HYPES in a corresponding P-HYPESs to train and test models, is often not feasible. The key point of our RGS strategy is to randomly generate a certain number of HYPESs from the correspond P-HYPESs following a uniform distribution, so that the grid search can be finished within a reasonable timeframe and every HYPES in the pool has an equal chance to be picked. RGS_HSG implements this strategy; 4) A grid searcher, which was developed based on the grid search libraries provided by Scikit_Learn. The grid searcher goes through every HYPES generated by RGS_HSG, and at each HYPES, it calls the DFNN model learner to train and test models using the current HYPES, and records the output information such as the current HYPES and the corresponding model performance scores.

In this study, we learned DNM models from the LSM datasets using RSGP. Specifically, we learned the DNM-5Year models from the LSM-5Year dataset, the DNM-10Year models from the LSM-10Year dataset, and the DNM-15Year models from the LSM-15Year dataset. These models are the all-feature models; Similarly, we learned the DNM_RF-5Year models from the LSM_RF-5Year dataset, the DNM_RF-10Year models from the LSM_RF-10Year dataset, and the DNM_RF-15Year models from the LSM_RF-15Year dataset, and these models are the RF models. We describe in details the grid search experiments we conducted to develop these models in the "Experiments" subsection below. We can predict for a new patient the risk of 5-year BCM using the DNM-5Year and DNM_RF-5Year models, 10-year BCM using the DNM-10Year and DNM_RF-10Year models, and 15-year BCM using the DNM-15Year and DNM_RF-15Year models.

**The extended RGSP package for developing the comparison ML models.**
As stated in the Introduction, another main purpose of this study is to compare the DFNN-based models with a set of representative ML models that are not based on ANNs. We included the following ML methods in this study: *Naïve Bayes* (*NB*), a simplified Bayesian network (BN) model which normally only contains one parent node and a set of children leaf nodes [30,35,36]. In a basic NB model, there is an edge from the parent to each of the children. There are multiple types of NB classifiers included in the Scikit-learn libraries. We used the categorical NB in this study because our datasets only contain categorical data; *Logistic Regression* (*LR*), a supervised learning classification method, which is normally suitable for binary classification problems [36,37]. We included this method because our outcome feature is a binary variable; *Decision Tree* (*DT*), one of the most widely used machine learning methods. It contains a tree-like structure in which each internal node represents a test on a feature and each leaf node represents a class value [38]. It can be used for both classification and regression tasks; *Support Vector Machine* (*SVM*), a machine learning method that tries to identify a hyperplane that has maximum margin defined by support vectors [39,40]. SVM can be used for both regression and classification tasks, and it is widely applied in the later. We used the SVC version of the SVM in this study, which uses a linear hyperplane to separate the data points. We therefore use SVM and SVC as exchangeable terms in this paper; *The least absolute shrinkage and selection operator* (*LASSO*), is a regression-based classifier that can be used to conduct variable selection and regularization in order to enhance prediction performance [41]; *K-Nearest Neighbors* (*KNN*), a supervised machine learning method that can be used for both classification and regression tasks [42]. KNN predicts the class value of an new case by its *k* nearest neighboring data points. To do this, KNN assumes that cases with similar covariate values are near to each other [42]; *Random Forests* (*RaF*) is a typical bagging type of ensemble method, in which the trainer will randomly select a certain amount of sample data and create corresponding decision trees to form a random forest [43]; *Adaptive Boosting* (*ADB*) is a typical boosting type of ensemble method. Unlike the RaF model, where each DT is independent, the next learner of ADB will adjust its prediction work based on the result of the previous weak learner that tends to make incorrect predictions [44]. *eXtreme Gradient Boosting* (*XGB*), another well-known boosting type of ensemble learning. Unlike ADB, it uses gradient boosting, which is based on the difference between true and predicted values to improve model performance [45].

We used the libraries provided in Scikit-Learn [33,46] to implement these ML classifiers. Just like the deep learning method, each of these ML methods has a set of adjustable hyperparameters (see Table S5 in the supplement) that can be tuned to optimize prediction performance. We extended our RGSP to include the nine ML methods. As we did for the DFNN method, we conducted grid searches using RGSP to learn and optimize the all-feature models from the LSM datasets, and the RF models from the LSM_RF datasets for each of the nine ML methods. For the nine ML methods, the all-feature models are named with their short method names, and the RF models are named as the short method names concatenated with '_RF'. For example, the all-features models for the *Naïve Bayes* method are called NB models, and the RF models for the *Naïve Bayes* method are called NB_RF models.

**The adjustable hyperparameters and their value selection**
As previously described, the RGS_HSG component of RGSP takes as an input a preselected set of values for each of the adjustable hyperparameters, and produces for a grid search a certain number of HYPESs randomly selected from the P-HYPESs. The number of HYPES is another input parameter of RGS_HSG. The P-HYPESs are determined by the sets of input values, one for each of the adjustable hyperparameters. The general rules we used to select an input set of values for an adjustable hyperparameter are as follows: if a hyperparameter contains a fixed number of values, then give all of them to RGS_HSG ; if a hyperparameter has an infinite number of values, for example, when it is a continuous variable, we first identify the minimum and the maximum values of this hyperparameter by a package called *Single Hyperparameter Grid Searches* (*SHGS*) [47]. We then select all values between the minimum and maximum values, separated by a multiple of a very small value called step size. For example, if for a hyperparameter, the minimum value is

0.001, and maximum value is 0.3, and if we use a step size of 0.001, then the set of values that we choose for the hyperparameter would include a sequence of 299 different values, which are 0.001, 0.002, 0.003, … , 0.298, 0.299, and 0.3. We applied these rules to all ten ML methods involved in this study including deep learning. The adjustable hyperparameters and their input values that we used for the ten ML methods are shown in Table S5 in the supplement.

**Experiments**
To compare the RF models with corresponding all-feature models for each of the ten ML methods including deep learning, we discovered the best RF and all-feature model for predicting 5-year, 10-year, and 15-year BCM each respectively. This gives us 6 best models per method, and 60 best models in total for the ten methods. We followed the same experiment procedure to identify each pair of the best RF and all-feature model, which we describe below, using the development of the best DNM-5Year and DNM_RF-5Year model as an example.

**Step 1.** Call RGS-HSG to randomly generate 6000 HYPESs. The input set of values for each of the hyperparameters to RGS-HSG are shown in S5, and the input number of HYPESs to RGS-HSG is 6000; **Step 2**. Run the grid searcher of RGSP. The set of 6000 HYPESs generated in Step 1 is one of the input to the grid searcher. Another input to the grid searcher is the dataset, namely, the 80% of the LSM-5Year data that serves as the train-test set ( see "performance metrics") . The grid searcher will go through each of the 6000 HYPESs, and train and test models following the 5-fold CV mechanism (see "performance metrics") at each HYPES. At each of the 6000 settings, five different models are training and tested, performance scores regarding model training and testing and the corresponding HYPES are all recorded as part of the output of the grid searcher. During this step, 30000 DNM-5Year models are trained and tested; **Step 3**. At the end of the grid search, the grid searcher will select the best HYPES that is associated with the best average performance score among all 6000 HYPES. The top DNM-5Year model will then be developed by refitting the entire train-test set of the LSM-5Year data using the best HYEPS; **Step 4**. Repeat Step 2 and Step 3 to develop the best DNM_RF-5Year models, but using the LSM_RF-5Year instead of LSM-5Year dataset.

In order to ensure a fair comparison between a pair of best RF and all-feature model, we used the exactly same 6000 HYPESs generated in a same Step 1, while followed a separate Step 2 and 3 when developing the two best models. But a different Step 1 was conducted for each different dataset and ML method. To further ensure the fair comparison, we tried our best to conduct separate grid searches for developing the two under the same computer arrangement in terms of computers used, CPU cores used per computer, and RAM allocations. We distributed the computing work load among a small cluster of computers manually to achieve this.

**Performance metrics and the 5-fold CV**
**ROC_curve and AUC**: Our grid searches use an *AUC* score to measure the prediction performance of a model. AUC stands for *area under the curve*, which was originated from what's so called a *receiver operator characteristic curve* (*ROC_curve*) which plots the *true positive rate* (*TPR*) against the *false positive rate* (*FPR*) at each of the cutoff values, given a test dataset and a prediction model [48]. The TPRs and FPRs are calculated based on the set of true outcome values contained in the test dataset and the corresponding set of predicted outcome values obtained from the prediction model. An AUC score measures the discrimination performance of a model.

**The 5-fold CV procedure, mean-test AUC, and validation AUC**: Our grid searchers follow the 5-fold CV mechanism to train and test models at each HYPES. In order to conduct a 5-fold CV in a grid search, we first split the dataset that will be using in the grid search. In general, we use the following procedure to split a dataset: 1) split the entire dataset into a train-test set that contains 80% of the cases and a validation set that contains 20% of the cases. The train-test set will be given to a grid search as the input dataset, and the validation set will be kept aside for later validation tests; 2) divide a train-test set evenly into 5 portions for the purpose of conducting a 5-fold CV. The division is mostly done randomly except that each portion should have approximately 20% of the positive cases and the negative cases respectively to ensure that it is a representative fraction of the dataset. During a 5-fold CV, five different models are generated and tested, each is trained using a unique combination of 4 portions of the train-test set, and tested using the remaining portion. Five AUC scores are produced based on the tests done by the five models, and the average of these scores, called *mean-test AUC*, is also computed and recorded. The best HYPES selection done at the end of a grid search is based on the mean-test AUC scores recorded, and the best model is developed by refitting the entire train-test set used by the grid search at the best HYPES. In this study, a ROC_curve for a selected prediction model is generated by testing the cases contained in the corresponding validation set using the model, and we call the AUC obtained from such a curve a *validation AUC*.

**The SHAP method for the explanation of a prediction**
The Shapley value was introduced by Lloyd Shapley in 1951. It represents the distribution of individual payoffs in cooperative games by measuring the marginal contribution of an individual to the collective outcome [49,50]. The *Shapley additive explanations* (*SHAP*), developed based the Shapley value, is a method that can be used to explain the predictive output of a machine learning model [51,52]. A SHAP value shows the importance of a feature for contributing to the predicted outcome value. How a SHAP value is computed can be explained using the following formula [51]:

$$\varphi_i(p) = \frac{1}{|F|} \sum_{S \subseteq F \setminus \{i\}} \frac{[p(S \cup \{i\}) - p(S)]}{\binom{|F|-1}{|S|}}$$

Each additive term of this formula has two components :1) the marginal contribution of the *i*th feature to the model's prediction; 2) the weight associated with the marginal contribution. $F$ represents the complete set of all features contained in the data, $i$ means the *i*th feature, for which we are computing the SHAP value, and $S$ represents a subset of $F$, which excludes the *i*th feature. $p(S \cup \{i\})$ represents the model's predicted outcome value using the combined set of features in S and $\{i\}$ as the predictors, while $p(S)$ represents the model's predicted outcome value using only the features in S as predictors. $p(S \cup \{i\}) - p(S)$ represents the contribution to the model's prediction, made by adding the *i*th feature to the subset S. For each subset, $1/(|F| \binom{|F|-1}{|S|})$ is given as the weight, determined by $|F|$, the size of the complete set of features F, and $|S|$, the size of S. The purpose of using the weight is to balance the overall prior influence among all possible sizes of S.

In this study, we use the Kernel Explainer of the SHAP library [51] to conduct SHAP analyses concerning the 60 best prediction models we obtained for the ten ML methods. To compute SHAP values, we first identify $k$ representative cluster centroids from a train-test set using the $k$-means clustering method. We then obtain the *background values* of features by computing the mean of the corresponding feature values of the $k$ centroids. In order to compute the SHAP value for the $i$th feature, the Kernel Explainer generates synthetic samples, and use them as the test cases for a model [51]. Each of the synthetic samples contains the true values from a validation set for a subset $S$ together with the background values for the remaining features in $F$ except for the $i$th feature. The $i$th feature assumes its true value from the validation set in a synthetic sample when the sample is used to obtain the $p(S \cup \{i\})$ from the model, and assumes its corresponding background value in a synthetic sample when it is used to obtain the $p(S)$ from the model. We generated SHAP bar plots, which show the mean absolute SHAP values of features from all test cases, and summary plots which show the SHAP value distributions of features among all test cases.

**RESULTS**

Table 2 below shows the side by side comparison of the best DNM and DNM_RF models in terms of their prediction performance measured by mean-test AUC for predicting 5-year, 10-year, and 15-year BCM, each respectively. We extended our experiments by conducting the same comparison for each of the nine non-deep learning ML methods, and the results are also included in Table 2. In addition, Table 2 contains a column named as "% Difference", which shows the percentage increase or decrease of the mean-test AUC of a best RF model from the corresponding best all-feature model. For example, for DNM models that predict 5-year BCM, the percentage difference is -2.09%, which means that the best DNM_RF-5Year model performs worse than the DNM-5Year model by 2.09% in terms of mean-test AUC.

As previously described in the Experiments section, each of the 60 best models that we obtained was developed based on the best HYPES that was selected from 6000 randomly picked HYPESs used in corresponding grid searches, based on the mean-test AUC of the five models trained at each HYPES. In addition to developing and comparing the best models obtained through grid searches, we are also interested in knowing and comparing the average performance of all models trained during corresponding grid searches. Due to this, we created Table 3 below, which contains, for each of the 60 best models, the average value of the 6000 mean-test AUCs associated with the 6000 HYPESs used in corresponding grid searches.

Each of the best models was obtained at the cost of training and testing tens of thousands of models through grid searches. A grid search with deep learning can be very time consuming. Table 4 below is a summary of the running time used by our grid searchers. We arrange these results in a side-by-side manner to compare the grid search time used for developing a best RF model with that used for developing the corresponding best all-feature model.

Another purpose of this study is to compare the prediction performance of the ten ML methods including deep learning. For predicting 5-year BCM, we compare in Figure 4(a) below the 20 best prediction models that we developed for the ten ML methods, including 10 best all-feature models and 10 best RF_models. We also rank the mean-test AUC scores of the 20 models from high to low, and show the rankings using a bar chart in Figure 4 (a). We did the same for the best 20 models for predicting 10-year BCM in Figure 4(b), and for predicting 15-year BCM in Figure 4 (c).

We also developed a ROC_curve for each of the 60 best models. Such a curve was developed by plotting FPRs against TPRs, obtained by testing cases in an independent validation set using a best model (see Methods for details). In Figure 5 below, we compare side by side the ROC_curves created for the best DNM and DNM_RF model concerning the risk prediction of 5-year (Figure 5(a)), 10-year (Figure 5(b)), and 15-year (Figure 5(c)) BCM, each respectively. Figure 6 compares the prediction performance of the ten ML methods in terms of their corresponding best models. It contains a panel of six subfigures. Each subfigure consists of 10 ROC_curves created for the corresponding best models of the ten ML methods. For example, Figure 6(a) contains the 10 curves for the best all-feature ML models concerning 5-year BCM.

We conducted SHAP analyses and developed a SHAP feature importance plot and summary plot for each of the 60 best prediction models. Due to page limit, we do not include all of the 120 plots in this paper, instead, we show side by side the SHAP bar charts of the two best DNM models concerning 15-year BCM in Figure 7, and the SHAP summary plots of these two models in Figure 8, as an example. We include in the supplement the SHAP summary plots for the two best DNM models concerning 5-year BCM and 10-Year BCM each respectively, and the two best models of each of the top three ML methods excluding DNM concerning 5-year, 10-year, and 15-year BCM each respectively. This gives us 22 plots included in 11 figures, namely, Figure S1 through Figure S11 in the supplement. The top ML methods were selected based on the rankings shown in Figure 4.

**Table2.** Side-by-side comparisons of the best all-feature and RF models of the ten ML methods

| 5 Year | | | | |
|---|---|---|---|---|
| Model | Mean-test AUC | Model | Mean-test AUC | % Difference |
| DNM | 0.766 | DNM_RF | 0.750 | -2.09% |
| ADB | 0.748 | ADB-RF | 0.695 | -7.09% |
| NB | 0.803 | NB_RF | 0.773 | -3.74% |
| DT | 0.756 | DT_RF | 0.740 | -2.12% |
| KNN | 0.747 | KNN_RF | 0.725 | -2.95% |
| LASSO | 0.772 | LASSO_RF | 0.719 | -6.87% |
| LR | 0.772 | LR_RF | 0.720 | -6.74% |
| RaF | 0.783 | RaF_RF | 0.755 | -3.58% |
| SVC | 0.725 | SVC_RF | 0.692 | -4.55% |
| XGB | 0.804 | XGB_RF | 0.762 | -5.22% |
| 10 Year | | | | |
| Model | Mean-test AUC | Model | Mean-test AUC | % Difference |
| DNM | 0.801 | DNM_RF | 0.782 | -2.37% |
| ADB | 0.763 | ADB-RF | 0.722 | -5.37% |
| NB | 0.801 | NB_RF | 0.785 | -2.00% |
| DT | 0.765 | DT_RF | 0.747 | -2.35% |
| KNN | 0.757 | KNN_RF | 0.746 | -1.45% |
| LASSO | 0.790 | LASSO_RF | 0.720 | -8.86% |
| LR | 0.790 | LR_RF | 0.722 | -8.61% |
| RaF | 0.806 | RaF_RF | 0.788 | -2.23% |
| SVC | 0.792 | SVC_RF | 0.758 | -4.29% |
| XGB | 0.811 | XGB_RF | 0.791 | -2.47% |

| 15 Year | | | | |
|---|---|---|---|---|
| Model | Mean-test AUC | Model | Mean-test AUC | Percent Difference |
| DNM | 0.818 | DNM_RF | 0.862 | 5.38% |
| ADB | 0.792 | ADB-RF | 0.840 | 6.06% |
| NB | 0.817 | NB_RF | 0.833 | 1.96% |
| DT | 0.800 | DT_RF | 0.786 | -1.75% |
| KNN | 0.782 | KNN_RF | 0.810 | 3.58% |
| LASSO | 0.804 | LASSO_RF | 0.847 | 5.35% |
| LR | 0.805 | LR_RF | 0.846 | 5.09% |
| RaF | 0.817 | RaF_RF | 0.815 | -0.24% |
| SVC | 0.807 | SVC_RF | 0.847 | 4.96% |
| XGB | 0.814 | XGB_RF | 0.829 | 1.84% |

**Table 3.** Mean-test AUC of the best model vs. average mean-test AUC of all corresponding models

| 5 Year | | | | | | |
|---|---|---|---|---|---|---|
| | Average 5 Year | Best 5 Year | % Difference | Average RF 5 Year | Best RF 5 Year | % Difference |
| DNM | 0.542 | 0.766 | 41.3% | 0.535 | 0.75 | 40.2% |
| ADB | 0.611 | 0.748 | 22.4% | 0.629 | 0.695 | 10.5% |
| NB | 0.78 | 0.803 | 2.9% | 0.756 | 0.773 | 2.2% |
| DT | 0.56 | 0.756 | 35.0% | 0.553 | 0.74 | 33.8% |
| KNN | 0.727 | 0.747 | 2.8% | 0.704 | 0.725 | 3.0% |
| LASSO | 0.77 | 0.772 | 0.3% | 0.717 | 0.719 | 0.3% |
| LR | 0.77 | 0.772 | 0.3% | 0.716 | 0.72 | 0.6% |
| RaD | 0.556 | 0.783 | 40.8% | 0.547 | 0.755 | 38.0% |
| SVC | 0.563 | 0.725 | 28.8% | 0.517 | 0.692 | 33.8% |
| XGBoost | 0.594 | 0.804 | 35.4% | 0.585 | 0.762 | 30.3% |
| 10 Year | | | | | | |
| | Average 10 Year | Best 10 Year | % Difference | Average RF 10 Year | Best RF 10 Year | % Difference |
| DNM | 0.553 | 0.801 | 44.8% | 0.553 | 0.782 | 41.4% |
| ADB | 0.692 | 0.763 | 10.3% | 0.691 | 0.722 | 4.5% |
| NB | 0.794 | 0.801 | 0.9% | 0.768 | 0.785 | 2.2% |
| DT | 0.556 | 0.765 | 37.6% | 0.544 | 0.747 | 37.3% |
| KNN | 0.744 | 0.757 | 1.7% | 0.728 | 0.746 | 2.5% |
| LASSO | 0.787 | 0.79 | 0.4% | 0.718 | 0.72 | 0.3% |
| LR | 0.787 | 0.79 | 0.4% | 0.718 | 0.722 | 0.6% |
| RaD | 0.549 | 0.806 | 46.8% | 0.544 | 0.788 | 44.9% |
| SVC | 0.605 | 0.792 | 30.9% | 0.571 | 0.758 | 32.7% |
| XGBoost | 0.569 | 0.811 | 42.5% | 0.559 | 0.791 | 41.5% |
| 15 Year | | | | | | |
| | Average 15 Year | Best 15 Year | % Difference | Average RF 15 Year | Best RF 15 Year | % Difference |
| DNM | 0.563 | 0.818 | 45.3% | 0.577 | 0.862 | 49.4% |
| ADB | 0.694 | 0.792 | 14.1% | 0.761 | 0.84 | 10.4% |
| NB | 0.785 | 0.817 | 4.1% | 0.764 | 0.833 | 9.0% |
| DT | 0.576 | 0.8 | 38.9% | 0.59 | 0.786 | 33.2% |
| KNN | 0.759 | 0.782 | 3.0% | 0.767 | 0.81 | 5.6% |
| LASSO | 0.799 | 0.804 | 0.6% | 0.843 | 0.847 | 0.5% |
| LR | 0.799 | 0.805 | 0.8% | 0.843 | 0.846 | 0.4% |
| RaD | 0.56 | 0.817 | 45.9% | 0.544 | 0.815 | 49.8% |
| SVC | 0.639 | 0.807 | 26.3% | 0.676 | 0.847 | 25.3% |
| XGBoost | 0.523 | 0.814 | 55.6% | 0.518 | 0.829 | 60.0% |

**Table 4.** Comparison of grid Search running time (in minutes) between corresponding all-feature and RF models

| Model | 5 Year | RF 5 Year | 10 Year | RF 10 Year | 15 Year | RF 15 Year |
|---|---|---|---|---|---|---|
| DNM | 85932.83 | 85402.63 | 31053.83 | 31110.23 | 6308.75 | 6215.888 |
| ADB | 40.85 | 1798.62 | 15.55 | 1971.38 | 10.33 | 563.555 |
| NB | 0.6 | 0.45 | 0.47 | 0.3 | 0.32 | 0.23 |
| DT | 0.33 | 0.28 | 0.22 | 0.17 | 0.18 | 0.13 |
| KNN | 22.77 | 20.17 | 5.15 | 4.37 | 1.4 | 1.28 |
| LASSO | 1.12 | 0.7 | 0.78 | 0.33 | 0.57 | 0.25 |
| LR | 0.92 | 0.58 | 0.8 | 0.38 | 0.63 | 0.28 |
| RaF | 25.65 | 25.32 | 20.73 | 15.3 | 17.53 | 16.35 |
| SVC | 9.7 | 8.58 | 4.02 | 3.82 | 0.9 | 0.68 |
| XGB | 26.08 | 19.3 | 14.58 | 10.55 | 14.40 | 12.58 |

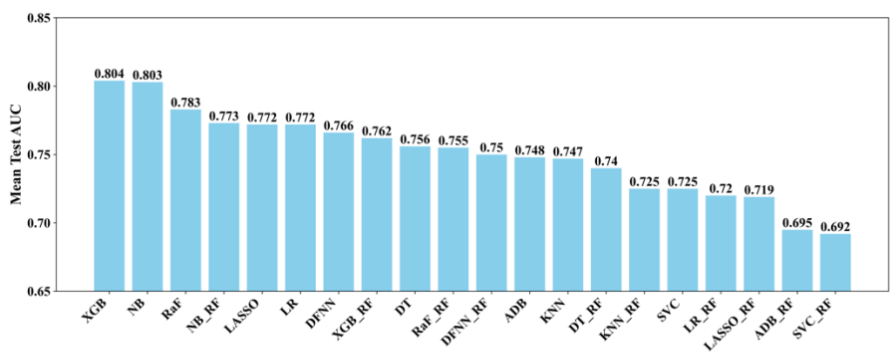

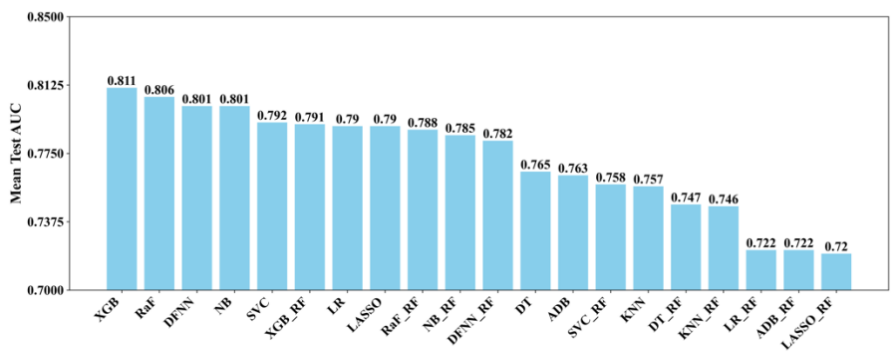

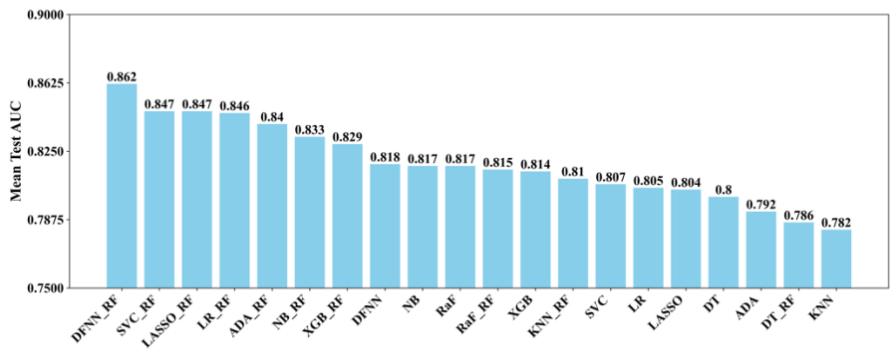

**Figure 4.** The rankings of the 20 best ML models shown in a bar chart: (a) 5-year; (b) 10-year; (c) 15-year

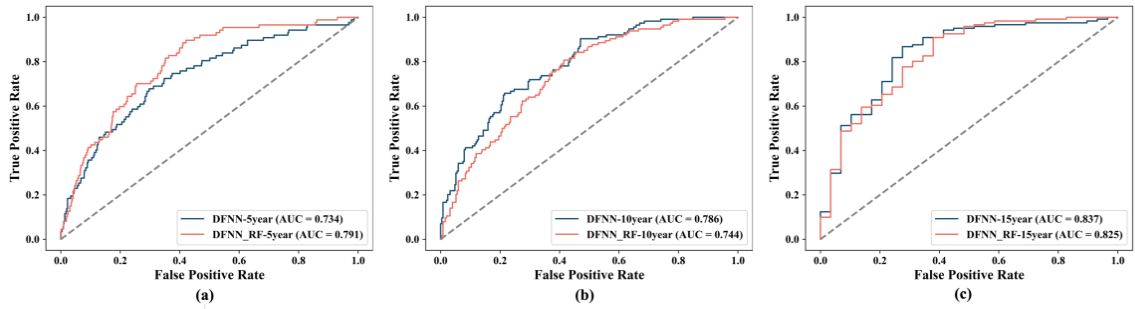

**Figure 5.** Side-by-side comparison of the ROC_curves for the best DNM and DNM_RF model concerning predicting 5-year ((a)), 10-year ((b)), and 15-year ((c)) BCM

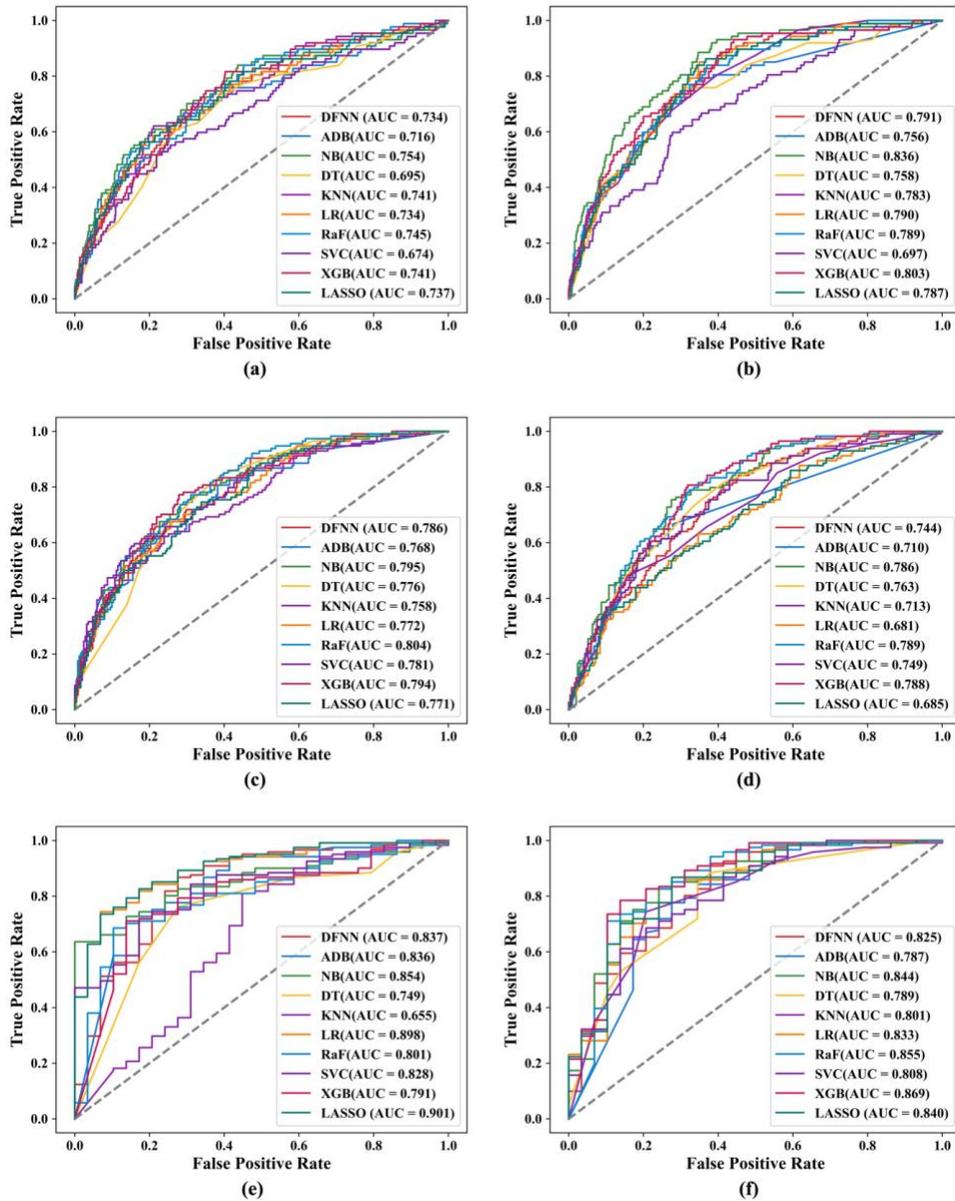

**Figure 6**. ROC_Curves for the best models of the ten ML Methods: (a) all-feature 5-year; (b) RF 5-year; (c) all-feature 10-year; (d) RF 10-year; (e) all-feature 15-year; (f) RF 15-year

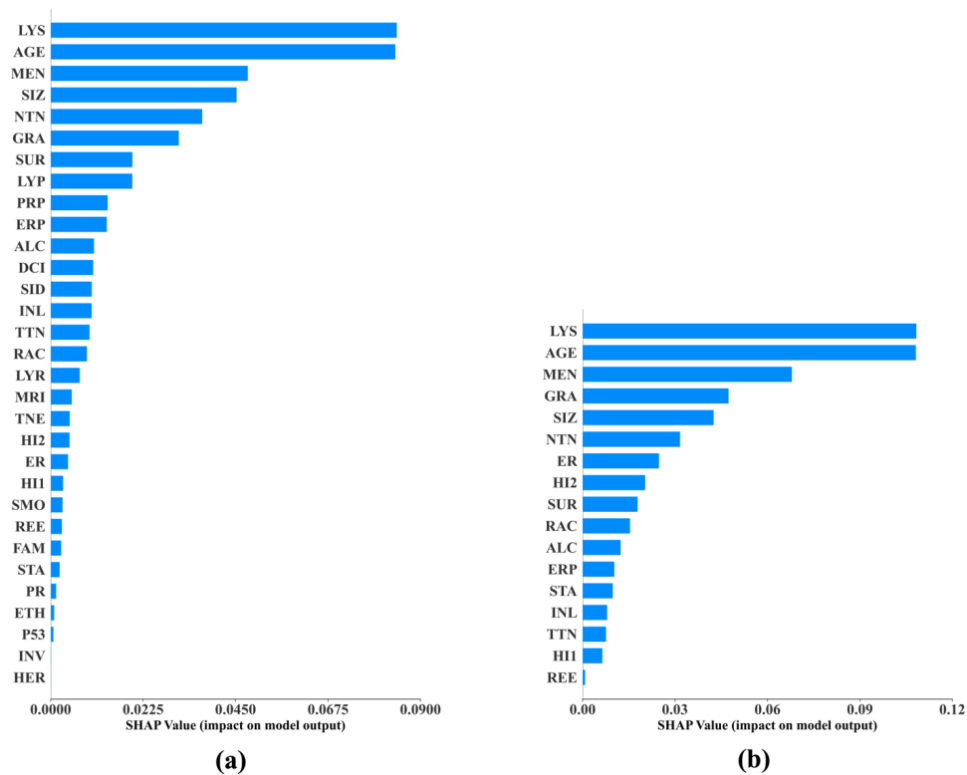

**Figure 7.** The SHAP bar plots for the best DNM_15-Year model ((a)), and DNM_RF-15Year model ((b))

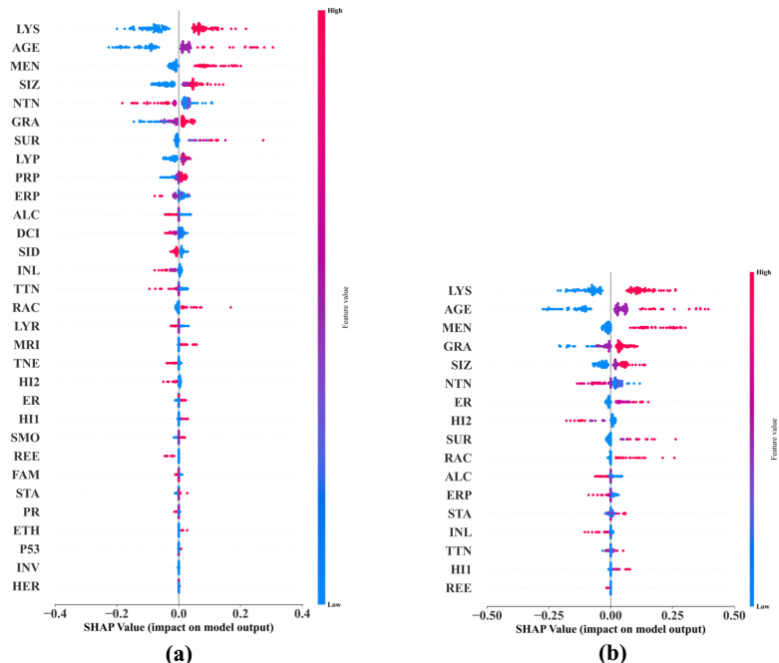

**Figure 8.** The SHAP Summary Plots for the best DNM-15Year model ((a)) and DNM_RF-15Year mode ((b))

Acronyms for Figure 7 and 8, **AGE**: age at diagnosis of the disease; **ALC**: alcohol usage; **DCI**: type of ductal carcinoma in situ; **ER**: estrogen receptor expression; **ERP**: percent of cell stain pos for ER receptors; **ETH**: ethnicity; **FAM**: family history of cancer; **GRA**: grade of disease; **HER**: HER2 expression; **HI1**: tumor histology; **HI2**: tumor histology subtypes; **INL**: where invasive tumor is located; **INV**: whether tumor is invasive; **LYP**: number of positive lymph nodes; **LYR**: number of lymph nodes

*removed;* ***LYS****: patient had any positive lymph nodes;* ***MEN****: inferred menopausal status;* ***MRI****: MRIs within 60 days of surgery;* ***NTN****: number of nearby cancerous lymph nodes;* ***PR****: progesterone receptor expression;* ***PRP****: percent of cell stain pos for PR receptors;* ***P53****: whether P53 is mutated;* ***RAC****: race;* ***REE****: removal of an additional margin of tissue;* ***SID****: side of tumor;* ***SIZ****: size of tumor in mm;* ***SMO****: smoking;* ***STA****: composite of size and # positive nodes;* ***SUR****: whether residual tumor;* ***TNE****: triple negative status in terms of patient being ER, PR, and HER2 negative;* ***TTN****: prime tumor stage in TNM system.*

## DISCUSSION

One of the main purposes of this study is to compare the DNM_RF models with the DNM models, and we assumed that the DNM_RF models should perform better. Our experiments demonstrate that this is indeed true in certain situations. Specifically, as shown in Table 2, when predicting 15-year BCM, the DNM_RF-15Year model (with a mean-test AUC of 0.862) performs about 5.4% better than the DNM_15Year model (with a mean-test AUC of 0.818). In addition, according to Figure 5(a), when predicting 5-year BCM, the DNM_RF-5Year model (with a validation AUC of 0.792) performs almost 8% better than the DNM-5Year model (with a validation AUC of 0.734). As described in the *Performance Metrics* subsection, a validation AUC is obtained by making predictions for the cases in the 20% data that were saved prior to grid searches, that is, these cases do not participate in any model training process. Therefore, a validation AUC reflects a model's capability of making correct predictions for new patients it has never seen. Thus, Figure 5(a) reveals that the DNM_ RF-5Year model is 8% more capable than the DNM-5Year model when dealing with new patients. Finally, if we consider all of the 60 best prediction models we obtained for the ten ML methods including deep learning, the DNM_RF-15Year is the best of the best by producing the highest mean-test AUC we have seen in our study. As shown by the bar chart in Figure 4(c), when predicting the risk of 15-year BCM, the DNM_RF model performs significantly better than all other best ML models including the best DNM model.

Recall that the DNM_RF-15Year model uses only a fraction of the features used by the DNM-15Year model, but it beats the latter when predicting the risk of 15-year BCM. This demonstrates the power of a BN-based method like MBIL in fishing for risk factors that are critical to a prediction. In addition, the results reveal the potential strength of a collaborative effort from different AI-based approaches such as deep learning and Bayesian networks. As we know, DNM_RF-15Year that scored the highest among all best models is a model resulted from a coalition of deep learning, Bayesian network, and grid search. The DNM_RF-5Year model that outperforms the DNM-5Year by 8% when conducting predictions for new patients is also a result of such an collaborative effort.

Although a RF model outperforms its corresponding non RF model in eight out of the ten ML methods when predicting 15-year BCM, as demonstrated by Table 2 and Figure 4, similar results are not seen for models that predict the risks of 5-year and 10-year BCM. We reckon this may somewhat further indicate the power of combing a ML method with grid search, since a all-feature model is a result of the coalition of these two. Recall we assumed that a RF model should beat its corresponding all-feature model mainly because the nonpredictive features remaining in the all-feature model may become noisy and therefore tamper the prediction capability of it. The results regarding 5-year and 10-year BCM may just tell us that the expected "noisy" effects have never occurred or been offset by something else, such as the coalition with a grid search. Without the "noisy" effects to take place, a good all-feature model should indeed be at least no worse than the RF one, as seen in the 5-year and 10 year cases, because all the good predictors in the latter are also available in the former, while some weak predictors that can be overlooked by a risk factor learner such as MBIL would only be available in the all-feature model.

The average mean-test AUC of all corresponding models trained, as seen in Table 3, shows the expected model prediction performance when the HYPESs of a grid search are randomly picked from the P-HYPESs. This should also be the expected model performance when the HYPES of a model is randomly selected from P-HYPES without doing a grid search. Based on Table 3, five out of the ten ML methods, including XGBoost, DFNN, RaF, DT, and SVM, benefit greatly from grid searches, with a percentage performance increase, from that of the best model discovered through grid searches to the expected performance at random without grid searches, ranging from 25.3% (with SVC_RF-15Year) up to 60% (with XGB_RF-15Year). Using the DNM_RF-15Year models as an example, the average mean-test AUC of all DNM_RF-15Year models trained is 0.577, while the mean-test AUC of the best DNM-15Year model discovered through grid searches is 0.862, that is, grid searches brought in an 49.4% performance increase in this case. DFNN is developed based on ANN, which can be considered as a special case of DFNN in that an ANN is just a DFNN that contains only one hidden layer. Both deep learning and grid search became popular long after ANN was first introduced. The results in Table 3 disclose that a Neural Network-based method is more sensitive to hyperparameter tunning than some other ML methods such as NB, LR, LASSO, KNN, and SVM, therefore help explain why ANN often performed worse than some other ML methods in earlier years since its invention when grid search was not applied [53–55].

We notice from Table 3 that some of the ML methods including LASSO, LR, NB, KNN, and ADB are less affected by grid search. The percentage performance increase of these methods ranges from 0.3% (LASSO) to 10.5% (ADB). LASSO and LR are the two methods least influenced by grid search, with less than 1% performance improvement across all their models. Although these five methods are less sensitive to grid search, some of them are among the top performers. For examples, LASSO_RF-15Year ranks number three out of all 15-year models, and LASSO-5Year ranks number five out of all 5-year models; NB-5Year ranks number two and NB_RF-5Year four among all 5-year models, and NB-10Year ranks number four among all 10-year models. In addition, NB-5Year and NB_RF-5Year both rank number among their own peers based on the validation AUCs as shown in Figure 6, indicating NB tends to do well when handling new patients. Therefore, when a grid search is not feasible, methods like LASSO and NB should not be bad choices in a similar prediction task.

It is not surprising to see that deep learning performs very well among all the ML methods, with DNM_RF-15Year ranking number one among all 15-year models, and DNM-10Year ranking number three among all 10 year models. It is worth mentioning that some of the other ML methods demonstrate good performance also. For example, the two ensemble methods, XGB and RaF are both top performers. XGB ranks number one among all other methods when predicting 5-year or 10-year BCM. RaF-5Year ranks number three, RaF-10Year two, and RaF-RF-15Year three, among its peers. These two methods also consistently excel when making predictions for new patients. As shown in Figure 6, which is created by using the validation AUCs, both RaF-10Year and RaF_RF-10Year rank number one among their own peers, and XGB_RF-15Year ranks the highest among its peers. Although DFNN, XGB, and RaF all benefit greatly from grid search, XGB and RaF use way less time than DFNN. For example, based on Table 4, it takes about 6216,

13, and 16 minutes to train DNM_RF-15, XGB_RF-15, and RaF_RF-15, with 6000 HYPESs each respectively. So when we have very limited time and budget to run grid searches, XGB and RaF are good alternatives for a costly deep learning.

The SHAP analyses reveal the relative importance of the predictors. Based on Figure 7, DNM and the DNM-RF agree completely that the top three most important features for predicting 15-year BCM are *lymph_node _status*, *age_at_diagnosis*, and *menopausal_status*. From Figure 8, we notice that the least important features such as *HER2*, *Invasive*, and *p53* found in the DNM-15Year model were not even included as predictors in the DNM_RF-15Year model. Recall that the DNM_15Year model was trained purely by the deep learning method, while the predictor inclusion of the DNM_RF-15Year model was solely determined by the BN-based MBIL method, indicating the least important features found by deep learning were also independently identified by MBIL. Therefore, DFNN and MBIL support the validity of each other in this regard.

**CONCLUSIONS**

This study demonstrates that deep learning, BN based methods, grid search are all powerful machine learning tools. Through the coalition of the three, we obtained the best mean_test AUC out of 1,800,000+ ML models that were trained and tested. The DNM_RF model, resulted from this coalition, outperformed all other ML models when predicting the risks of 15-year BCM. The effectiveness of the BN-based MBIL method in identifying risk factors for a disease outcome is further substantiated through this study. The grid search mechanism is shown to be a very powerful prediction optimization method not only for deep learning but also for some of the other ML methods. Surprisingly, some of the ML methods such as the ensemble XGB and *Random Forrests* can outperform deep learning through grid searches, but take "no time" relative to deep learning, while some other ML methods such as BN-based Naïve Bayes, LASSO, LR, and KNN, overall less sensitive to grid search, can sometimes excel in prediction also. This seems to indicate that these non-deep learning ML methods, sensitive or not to grid search, are sufficiently good alternatives for deep learning in conducting similar prediction tasks, especially when budget is low and time is limited.

**DECLARATIONS**

**Ethics approval and consent to participate**
The study was approved by University of Pittsburgh Institutional Review Board (IRB # 196003) and the U.S. Army Human Research Protection Office (HRPO # E01058.1a).
The need for patient consent was waived by the ethics committees because the data consists only of de-identified data that are publicly available.

**Consent for publication**
Not applicable.

**Availability of data and material**
The data used in this study are available at datadryad.org (DOI 10. 5061/dryad.64964m0).

**Competing interests**

The authors declare that they have no competing interests.

**Funding**
Research reported in this paper was supported by the U.S. Department of Defense through the Breast Cancer Research Program under Award No. W81XWH-19-1-0495 (to XJ). Other than supplying funds, the funding agencies played no role in the research.

**Authors' Contribution**
XJ originated the study and designed the methods. XJ and YZ wrote the first draft of the manuscript, implemented the methods, and prepared and analyzed the results. YZ conducted experiments. All authors contributed to the preparation and revision of the manuscript. All work was conducted in the University of Pittsburgh.

# Supplement

**Table S1.** The variables of the LSM datasets

| Abbreviation | Variable Name | Description | Values |
|---|---|---|---|
| RAC | *race* | race of patient | white, black, Asian, American Indian or Alaskan native, native Hawaiian or other Pacific islander |
| ETH | *ethnicity* | ethnicity of patient | not Hispanic, Hispanic |
| SMO | *smoking* | smoking history of patient | ex smoker, non smoker, cigarettes, chewing tobacco, cigar |
| ALC | *alcohol usage* | alcohol usage of patient | moderate, no use, use but nos (non otherwise specified), former user, heavy user |
| FAM | *family history* | family history of cancer | cancer, no cancer, breast cancer, other cancer, cancer but nos |
| AGE | *age_at_diagnosis* | age at diagnosis of the disease | 0-49, 50-69, >69 |
| MEN | *menopausal_status* | inferred menopausal status | pre, post |
| SID | *side* | side of tumor | left, right |
| TNE | *TNEG* | triple negative status in terms of patient being ER, PR, and HER2 negative | yes, no |
| ER | *ER* | estrogen receptor expression | neg, pos, low pos |
| ERP | *ER_percent* | percent of cell stain pos for ER receptors | 0-20, 20-90, 90-100 |
| PR | *PR* | progesterone receptor expression | neg, pos, low pos |
| PRP | *PR_percent* | percent of cell stain pos for PR receptors | 0-20, 20-90, 90-100 |
| P53 | *P53* | whether P53 is mutated | neg, pos, low pos |
| HER | *HER2* | HER2 expression | neg, pos |
| TTN | *t_tnm_stage* | prime tumor stage in TNM system | 0, 1, 2, 3, 4, IS, 1mic, X |
| NTN | *n_tnm_stage* | # of nearby cancerous lymph nodes | 0, 1, 2, 3, 4, X |
| STA | *stage* | composite of size and # positive nodes | 0, 1, 2, 3 |
| LYR | *lymph_nodes_removed* | number of lymph nodes removed | 0-11, 12-22, > 22 |
| LYP | *lymph_nodes_positive* | number of positive lymph nodes | 0, 1-8, >8 |
| LYS | *lymph_node_status* | patient had any positive lymph nodes | neg, pos |
| HI1 | *histology* | tumor histology | lobular, duct |
| SIZ | *size* | size of tumor in mm | 0-32, 32-70, >70 |
| GRA | *grade* | grade of disease | 1, 2, 3 |
| INV | *invasive* | whether tumor is invasive | yes, no |
| HI2 | *histology2* | tumor histology subtypes | IDC, DCIS, ILC, NC |
| INL | *invasive_tumor_location* | where invasive tumor is located | mixed duct and lobular, duct, lobular, none |
| DCI | *DCIS_level* | type of ductal carcinoma in situ | solid, apocrine, cribriform, dcis, comedo, papillary, micropapillary |
| REE | *re_excision* | removal of an additional margin of tissue | yes, no |
| SUR | *surgical_margins* | whether residual tumor | res. tumor, no res. tumor, |

| | | | no primary site surgery |
|---|---|---|---|
| MRI | *MRIs_60_surgery* | MRIs within 60 days of surgery | yes, no |
| | Metastasis | This is the outcome variable | yes, no |

**Table S2.** Predictors in the LSM_RF-5Year Dataset

| | Predictors | Description | Values |
|---|---|---|---|
| 1 | *race* | *race of patient* | *white, black, Asian, American Indian or Alaskan native, native Hawaiian or other Pacific islander* |
| 2 | *smoking* | smoking history of patient | ex smoker, non smoker, cigarettes, chewing tobacco, cigar |
| 3 | *family history* | family history of cancer | cancer, no cancer, breast cancer, other cancer, cancer but nos |
| 4 | *age_at_diagnosis* | age at diagnosis of the disease | 0-49, 50-69, >69 |
| 5 | *TNEG* | triple negative status in terms of patient being ER, PR, and HER2 negative | yes, no |
| 6 | *ER* | estrogen receptor expression | neg, pos, low pos |
| 7 | *ER_percent* | percent of cell stain pos for ER receptors | 0-20, 20-90, 90-100 |
| 8 | *PR* | progesterone receptor expression | neg, pos, low pos |
| 9 | *PR_percent* | percent of cell stain pos for PR receptors | 0-20, 20-90, 90-100 |
| 10 | *P53* | *P53* | *whether P53 is mutated* |
| 11 | *HER2* | HER2 expression | neg, pos |
| 12 | *t_tnm_stage* | *prime tumor stage in TNM system* | *0, 1, 2, 3, 4, IS, 1mic, X* |
| 13 | *n_tnm_stage* | *# of nearby cancerous lymph nodes* | *0, 1, 2, 3, 4, X* |
| 14 | *stage* | *composite of size and # positive nodes* | *0, 1, 2, 3* |
| 15 | *lymph_nodes_positive* | number of positive lymph nodes | 0, 1-8, >8 |
| 16 | *histology* | tumor histology | lobular, duct |
| 17 | *size* | *size of tumor in mm* | *0-32, 32-70, >70* |
| 18 | *invasive_tumor_location* | *where invasive tumor is located* | *mixed duct and lobular, duct, lobular, none* |
| 19 | *DCIS_level* | type of ductal carcinoma in situ | *solid, apocrine, cribriform, dcis, comedo, papillary, micropapillary* |
| 20 | *surgical_margins* | whether residual tumor | res. tumor, no res. tumor, no primary site surgery |

**Table S3.** Predictors in the LSM_RF-10Year Dataset

| | Predictors | Description | Values |
|---|---|---|---|
| 1 | *ethnicity* | ethnicity of patient | not Hispanic, Hispanic |
| 2 | *smoking* | smoking history of patient | ex smoker, non smoker, cigarettes, chewing tobacco, cigar |
| 3 | *alcohol usage* | alcohol usage of patient | moderate, no use, use but nos (non otherwise specified), former user, heavy user |
| 4 | *family history* | family history of cancer | cancer, no cancer, breast cancer, other cancer, cancer but nos |
| 5 | *age_at_diagnosis* | age at diagnosis of the disease | 0-49, 50-69, >69 |
| 6 | *TNEG* | triple negative status in terms of | yes, no |

|   |   | patient being ER, PR, and HER2 negative |   |
|---|---|---|---|
| 7 | *ER* | estrogen receptor expression | neg, pos, low pos |
| 8 | *ER_percent* | percent of cell stain pos for ER receptors | 0-20, 20-90, 90-100 |
| 9 | *PR* | progesterone receptor expression | neg, pos, low pos |
| 10 | *PR_percent* | percent of cell stain pos for PR receptors | 0-20, 20-90, 90-100 |
| 11 | *HER2* | HER2 expression | neg, pos |
| 12 | *n_tnm_stage* | # of nearby cancerous lymph nodes | 0, 1, 2, 3, 4, X |
| 13 | *stage* | composite of size and # positive nodes | 0, 1, 2, 3 |
| 14 | *lymph_nodes_positive* | number of positive lymph nodes | 0, 1-8, >8 |
| 15 | *histology* | tumor histology | lobular, duct |
| 16 | *grade* | grade of disease | 1, 2, 3 |
| 17 | *DCIS_level* | type of ductal carcinoma in situ | solid, apocrine, cribriform, dcis, comedo, papillary, micropapillary |
| 18 | *surgical_margins* | whether residual tumor | res. tumor, no res. tumor, no primary site surgery |

**Table S4.** Predictors in the LSM_RF-15 Year Dataset

|   | Predictors | Description | Values |
|---|---|---|---|
| 1 | *race* | race of patient | white, black, Asian, American Indian or Alaskan native, native Hawaiian or other Pacific islander |
| 2 | *alcohol usage* | alcohol usage of patient | moderate, no use, use but nos (non otherwise specified), former user, heavy user |
| 3 | *age_at_diagnosis* | age at diagnosis of the disease | 0-49, 50-69, >69 |
| 4 | *menopausal_status* | inferred menopausal status | pre, post |
| 5 | *ER* | estrogen receptor expression | neg, pos, low pos |
| 6 | *ER_percent* | percent of cell stain pos for ER receptors | 0-20, 20-90, 90-100 |
| 7 | *t_tnm_stage* | prime tumor stage in TNM system | 0, 1, 2, 3, 4, IS, 1mic, X |
| 8 | *n_tnm_stage* | # of nearby cancerous lymph nodes | 0, 1, 2, 3, 4, X |
| 9 | *stage* | composite of size and # positive nodes | 0, 1, 2, 3 |
| 10 | *lymph_node_status* | patient had any positive lymph nodes | neg, pos |
| 11 | *size* | size of tumor in mm | 0-32, 32-70, >70 |
| 12 | *grade* | grade of disease | 1, 2, 3 |
| 13 | *histology2* | tumor histology subtypes | IDC, DCIS, ILC, NC |
| 14 | *invasive_tumor_location* | where invasive tumor is located | mixed duct and lobular, duct, lobular, none |
| 15 | *re_excision* | removal of an additional margin of tissue | yes, no |
| 16 | *surgical_margins* | whether residual tumor | res. tumor, no res. tumor, no primary site surgery |
| 17 | *histology* | tumor histology | lobular, duct |

**Table S5.** The ML hyperparameters and their values given to the RGSP

| Method | Hyperparameter | Description | Values |
|---|---|---|---|
| DFNN | Epochs | Number of times model is trained by the full training dataset | 5 ~ 1001, step size 3 |
| | Batch size | Number of samples that are processed together in a single forward and backward pass during training | 1 to the # of datapoints in a dataset |
| | Learning rate | Control the learning and parameter update speed during optimization | 0.001 ~ 0.3, Step size 0.001 |
| | Dropout rate | Mitigate overfitting and training time by randomly ignoring nodes | 0 ~ 0.9, Step size 0.01 |
| | Momentum | Speed up optimization by incorporating historical gradients into parameter updates. Momentum is exclusively applicable within the SGD optimizer. | 0.1 ~ 0.9, Step size 0.01 |
| | decay | Iterative decay of the learning rate by applying a decreasing factor at each epoch | 0 ~ 0.3, Step size 0.001 |
| | L1 weight | Control the strength of L1 regularization in model training | 0 ~ 0.3, Step size 0.001 |
| | L2 weight | Control the strength of L2 regularization in model training | 0 ~ 0.3, Step size 0.001 |
| | # of Hidden Layers | The depth of a DNN model | 1,2,3,4 |
| | # of Hidden Nodes | Number of neurons in a hidden layer | All integers from 1 to the number of datapoints of each dataset |
| | Optimizer | Optimizes model parameters during training process towards minimizing the loss | SGD, Adam, Adagrad, Nadam, Adamax |
| | Weight Initializer | A technique employed to assign initial values to the weights of the connections between neurons in a neural network | Constant, Glorot_normal, Glorot_uniform, He_normal, He_uniform |
| | Input Layer Activation Function | A function applied to the input data of a neural network's input layer | Relu, Sigmoid, Softmax, Tanh |
| | Hidden Layer Activation Function | A function applied to the output of a hidden layer in a neural network | Relu, Sigmoid, Softmax, Tanh |
| | Output Layer Activation Function | A function applied to the output of a neural network's output layer, producing the final prediction of the network. | Sigmoid |
| | Loss Function | A method of evaluating the dissimilarity between the predicted output of a model and the actual values | Binary_crossentropy |
| AdaBoost | Base_estimator | The model that AdaBoost will enhance through iterative updates | Logistic Regression, Random Forest, Decision Tree |
| | n_estimators | The maximum number of estimators at which boosting is terminated | 1 ~ 1001, step size 1 |
| | Learning rate | The weight applied to each classifier at each boosting iteration | 0.001 ~ 0.101, step size 0.001 |
| | algorithm | The technique AdaBoost uses to update weights and make predictions | SAMME |

| Model | Hyperparameter | Description | Range |
|---|---|---|---|
| Naïve Bayes | alpha | The additive smoothing parameter | 0.00001 ~ 100, step size 0.00001 |
| | Min_categories | The minimum number of categories per feature | 5 ~ 16, step size 1 |
| | Fit_prior | Defines whether to learn class prior probabilities or not | True, false |
| Decision tree | Criterion | The function to measure the quality of a split | Gini, Entropy, Log_loss |
| | Splitter | The strategy used to choose the split at each node | Best, Random |
| | Max_depth | The maximum depth of the tree | 1 ~ 100, step size 1 |
| | Min_samples_split | The minimum number of samples required to split an internal node | 2 to the # of datapoints in a dataset, step size 1 |
| | Min_samples_leaf | The minimum number of samples required to be at a leaf node | 2 to half of the # of datapoints in a dataset, step size 1 |
| | Min_weight_fraction_leaf | The minimum weighted fraction of the sum total of weights | 0 ~ 0.5, step size 0.001 |
| | Max_features | The number of features to consider when looking for the best split | 1 ~ 17, step size 1 |
| | Max_leaf_nodes | The maximum number of leaf nodes a tree can have | 2 to the # of datapoints in a dataset, step size 1 |
| | Min_impurity_decrease | The hyperparameter controls the minimum reduction in impurity required for a split to be made at a node | 0 ~ 0.01, step size 0.001 |
| | Class_weight | The weights associated with classes in the form | Balanced, None |
| KNN | N_neighbors | Number of neighbors to use | 1 ~ 100, step size 1 |
| | Weights | The weight function used in prediction | Uniform, Distance, None |
| | Algorithm | The algorithm used to compute the nearest neighbors | Auto, Ball_tree, Kd_tree, Brute |
| | Leaf_size | The parameter that affects the speed of the construction and query | 1 to the # of datapoints in a dataset, step size 1 |
| LASSO | Max_iter | The maximum number of iterations | 5 ~ 1000, step size 1 |
| | Tol | The tolerance for the optimization | 0.00001 ~ 0.01, step size 0.00005 |
| | C | The inverse of regularization strength | 0 ~ 100, step size 0.00005 |
| | Class_weight | The parameter that adjusts the weights | Balanced, None |
| | Solver | The algorithm used by the optimization problem | Liblinear, Saga |
| Logistic Regression | Max_iter | The maximum number of iterations taken for the solvers to converge | 5 ~ 1000, step size 1 |
| | Tol | The tolerance for stopping criteria | 0.00001 ~ 0.01, step size 0.00005 |
| | C | The inverse of regularization strength | 0 ~ 100, step size 0.005 |
| | Class_weight | The weights associated with classes in the form | Balanced, none |
| | Solver | The algorithm that used in the optimization probem | None, L1, L2, Elasticnet |
| Random Forest | N_estimators | The number of tress in the forest | 1 ~ 1000, step size 1 |
| | Criterion | The function to measure the quality of a split | Gini, Entropy, Log_loss |
| | Max_depth | The maximum depth of the tree | 1 ~ 101, step size 1 |

|  | Min_samples_split | The minimum number of samples required to split an internal node | 2 to the # of datapoints in a dataset, step size 1 |
|  | Min_samples_leaf | The minimum number of samples | 1 to half of the # of datapoints in a dataset, step size 1 |
|  | Min_weight_fraction_leaf | The minimum weighted fraction of the sum total of weights | 0 ~ 0.5,step size 0.001 |
|  | Max_features | The number of features to consider when looking for the best split | 1 ~ 17,step size 1 |
|  | Max_leaf_nodes | The maximum number of leaf nodes a tree can have | 2 to half of the # of datapoints in a dataset, step size 1 |
|  | Min_impurity_decrease | The hyperparameter controls the minimum reduction in impurity required for a split to be made at a node | 0 ~ 0.01,step size 0.001 |
|  | Class_weight | The weights associated with classes in the form | Balanced, None |
| SVC | C | The regularization parameter | 0 ~ 100,step size 0.005 |
|  | Kernel | The hyperparameter that specifies the kernel type to be used in the algorithm | Linear, Poly, Rbf, Sigmoid |
|  | Degree | The degree of the polynomial kernel function | 0 ~ 5,step size 1 |
|  | Gamma | The kernel coefficient | Scale, Auto |
|  | Shrinking | Defines whether to use shrinking heuristic | True, false |
|  | Tol | The tolerance for stopping criterion | 0.00001 ~ 0.01,step size 0.00005 |
|  | Class_weight | The weights associated with classes in the form | Balanced, None |
|  | Max_iter | The maximum number of iterations taken for the solvers to converge | 5 ~ 1000, step size 1 |
| XGBoost | Booster | The hyperparameter defines which booster to use | Gbtree, Gblinear, Dart |
|  | Eta | The hyperparameter defines step size shrinkage for tree booster | 0.001 ~ 0.1,step size 0.001 |
|  | Gamma | The minimum loss reduction required to make a further partition on a leaf node of the tree for tree booster | 0 ~ 10, step size 0.01 |
|  | Max_depth | The maximum depth of a tree for tree booster | 1 ~ 100, step size 1 |
|  | Subsample | The hyperparameter defines the subsample ratio of the training instances | 0.01~ 1, step size 0.01 |
|  | Sampling_method | The method to use to sample the training instances | Uniform, Gradient_based |
|  | Alpha | The L1 regularization term on weights | 0 ~ 100,s tep size 0.00001 |
|  | Lambda | The L2 regularization term on weights | 0 ~ 100, step size 0.00001 |
|  | Tree_method | The tree construction algorithm used in XGBoost | Auto, Exact, Approx, Hist |
|  | Objective | The hyperparameter specifies the learning objective | Binary:logistic |

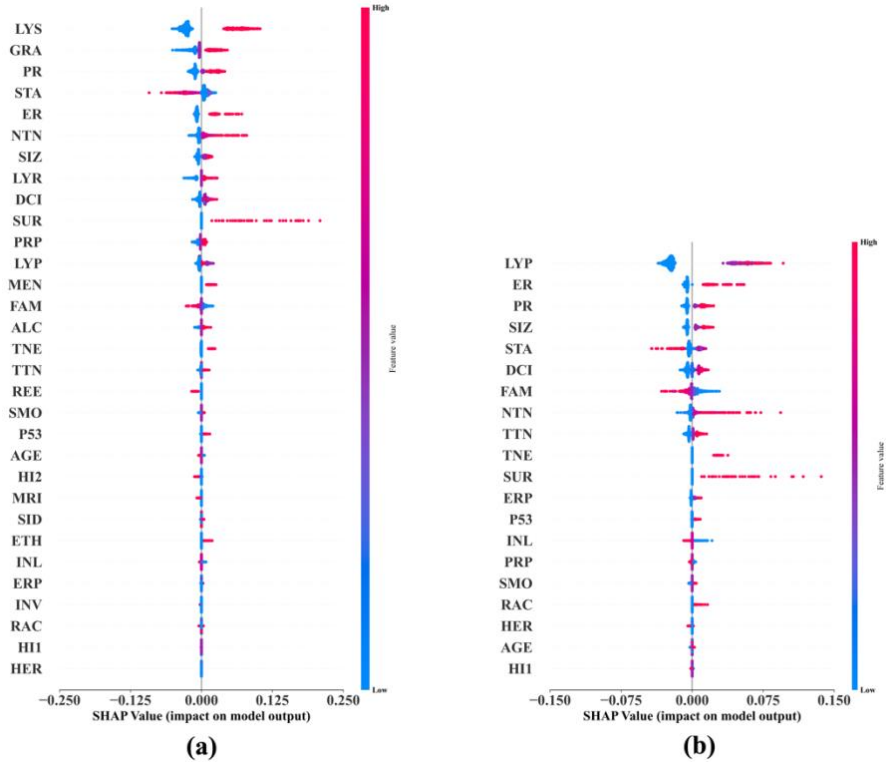

**Figure S1.** SHAP summary plots for the two best DFNN-based models concerning 5-year BCM:
(a) DNM-5Year; and (b) DNM_RF-5Year

*AGE: age at diagnosis of the disease; ALC: alcohol usage; DCI: type of ductal carcinoma in situ; ER: estrogen receptor expression; ERP: percent of cell stain pos for ER receptors; ETH: ethnicity; FAM: family history of cancer; GRA: grade of disease; HER: HER2 expression; HI1: tumor histology; HI2: tumor histology subtypes; INL: where invasive tumor is located; INV: whether tumor is invasive; LYP: number of positive lymph nodes; LYR: number of lymph nodes removed; LYS: patient had any positive lymph nodes; MEN: inferred menopausal status; MRI: MRIs within 60 days of surgery; NTN: number of nearby cancerous lymph nodes; PR: progesterone receptor expression; PRP: percent of cell stain pos for PR receptors; P53: whether P53 is mutated; RAC: race; REE: removal of an additional margin of tissue; SID: side of tumor; SIZ: size of tumor in mm; SMO: smoking; STA: composite of size and # positive nodes; SUR: whether residual tumor; TNE: triple negative status in terms of patient being ER, PR, and HER2 negative; TTN: prime tumor stage in TNM system.*

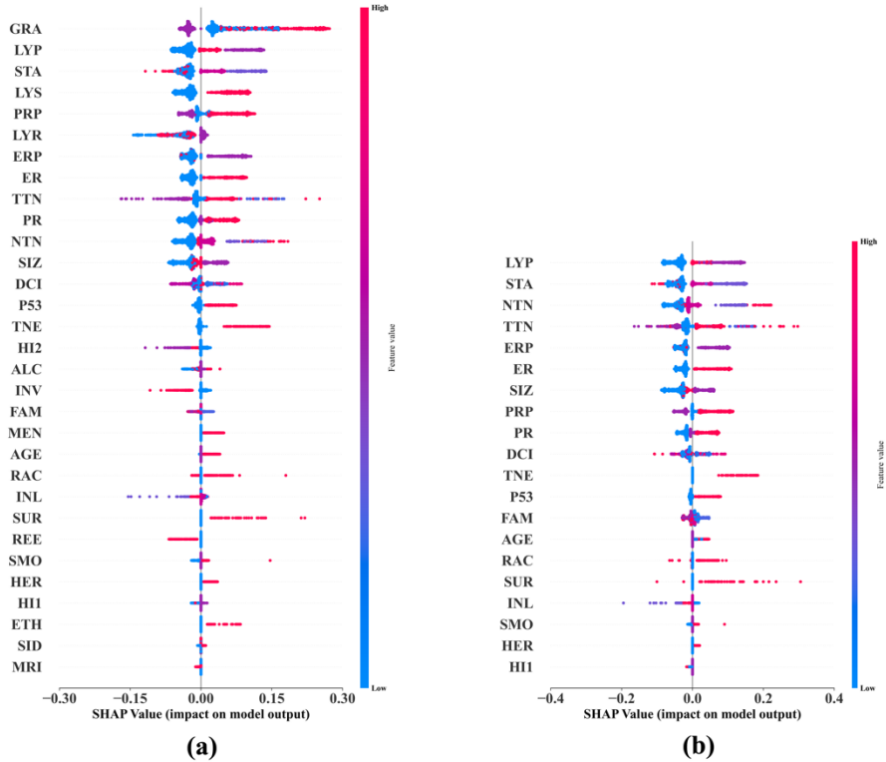

**Figure S2.** SHAP summary plots for the two best *Naïve Bayes* models concerning 5-year BCM: (a) NB-5Year; and (b) NB_RF-5Year

*AGE*: age at diagnosis of the disease; *ALC*: alcohol usage; *DCI*: type of ductal carcinoma in situ; *ER*: estrogen receptor expression; *ERP*: percent of cell stain pos for ER receptors; *ETH*: ethnicity; *FAM*: family history of cancer; *GRA*: grade of disease; *HER*: HER2 expression; *HI1*: tumor histology; *HI2*: tumor histology subtypes; *INL*: where invasive tumor is located; *INV*: whether tumor is invasive; *LYP*: number of positive lymph nodes; *LYR*: number of lymph nodes removed; *LYS*: patient had any positive lymph nodes; *MEN*: inferred menopausal status; *MRI*: MRIs within 60 days of surgery; *NTN*: number of nearby cancerous lymph nodes; *PR*: progesterone receptor expression; *PRP*: percent of cell stain pos for PR receptors; *P53*: whether P53 is mutated; *RAC*: race; *REE*: removal of an additional margin of tissue; *SID*: side of tumor; *SIZ*: size of tumor in mm; *SMO*: smoking; *STA*: composite of size and # positive nodes; *SUR*: whether residual tumor; *TNE*: triple negative status in terms of patient being ER, PR, and HER2 negative; *TTN*: prime tumor stage in TNM system.

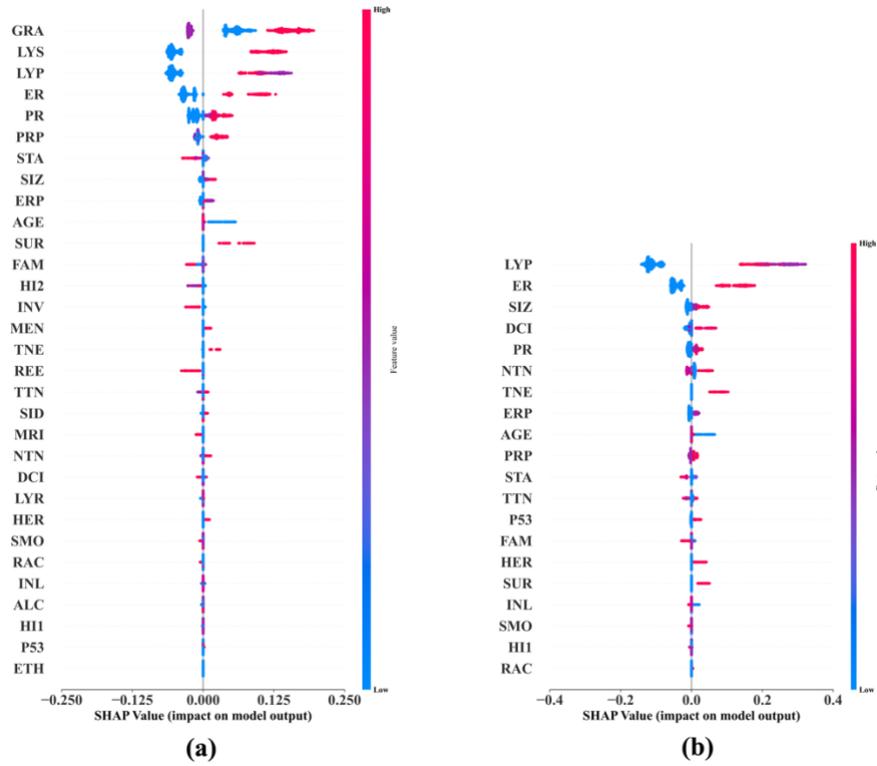

**Figure S3.** SHAP summary plots for the two best *Random Forests* models concerning 5-year BCM:
(a) RaF-5Year; and (b) RaF_RF-5Year

*AGE: age at diagnosis of the disease; ALC: alcohol usage; DCI: type of ductal carcinoma in situ; ER: estrogen receptor expression; ERP: percent of cell stain pos for ER receptors; ETH: ethnicity; FAM: family history of cancer; GRA: grade of disease; HER: HER2 expression; HI1: tumor histology; HI2: tumor histology subtypes; INL: where invasive tumor is located; INV: whether tumor is invasive; LYP: number of positive lymph nodes; LYR: number of lymph nodes removed; LYS: patient had any positive lymph nodes; MEN: inferred menopausal status; MRI: MRIs within 60 days of surgery; NTN: number of nearby cancerous lymph nodes; PR: progesterone receptor expression; PRP: percent of cell stain pos for PR receptors; P53: whether P53 is mutated; RAC: race; REE: removal of an additional margin of tissue; SID: side of tumor; SIZ: size of tumor in mm; SMO: smoking; STA: composite of size and # positive nodes; SUR: whether residual tumor; TNE: triple negative status in terms of patient being ER, PR, and HER2 negative; TTN: prime tumor stage in TNM system.*

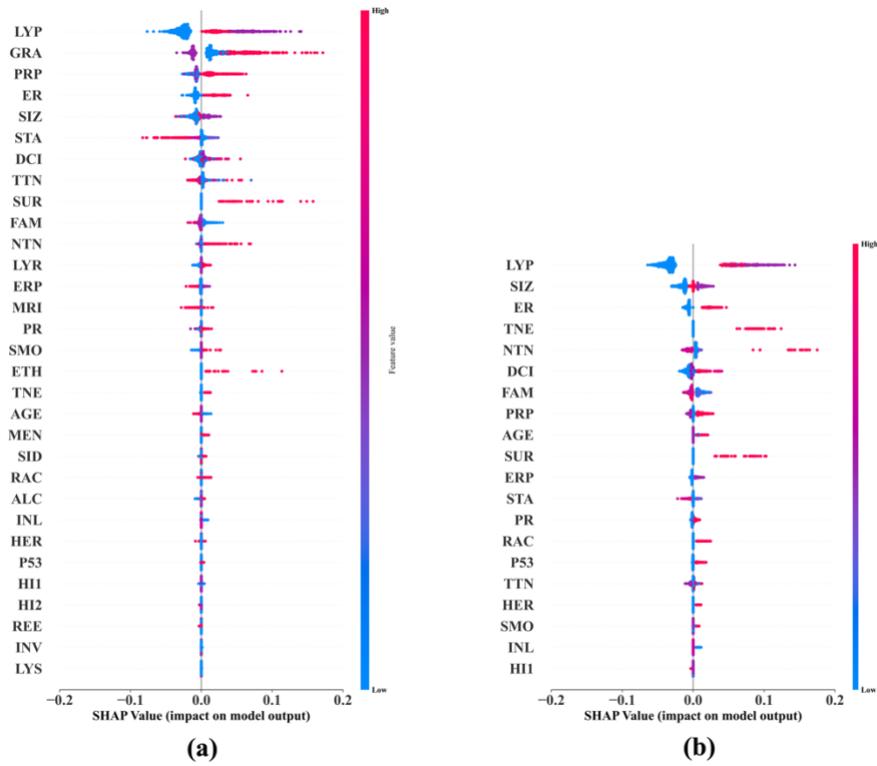

**Figure S4.** SHAP summary plots for the best two XGBoost Models concerning 5-year BCM
(a) XGB-5Year; and (b) XGB_RF-5Year

*AGE*: age at diagnosis of the disease; *ALC*: alcohol usage; *DCI*: type of ductal carcinoma in situ; *ER*: estrogen receptor expression; *ERP*: percent of cell stain pos for ER receptors; *ETH*: ethnicity; *FAM*: family history of cancer; *GRA*: grade of disease; *HER*: HER2 expression; *HI1*: tumor histology; *HI2*: tumor histology subtypes; *INL*: where invasive tumor is located; *INV*: whether tumor is invasive; *LYP*: number of positive lymph nodes; *LYR*: number of lymph nodes removed; *LYS*: patient had any positive lymph nodes; *MEN*: inferred menopausal status; *MRI*: MRIs within 60 days of surgery; *NTN*: number of nearby cancerous lymph nodes; *PR*: progesterone receptor expression; *PRP*: percent of cell stain pos for PR receptors; *P53*: whether P53 is mutated; *RAC*: race; *REE*: removal of an additional margin of tissue; *SID*: side of tumor; *SIZ*: size of tumor in mm; *SMO*: smoking; *STA*: composite of size and # positive nodes; *SUR*: whether residual tumor; *TNE*: triple negative status in terms of patient being ER, PR, and HER2 negative; *TTN*: prime tumor stage in TNM system.

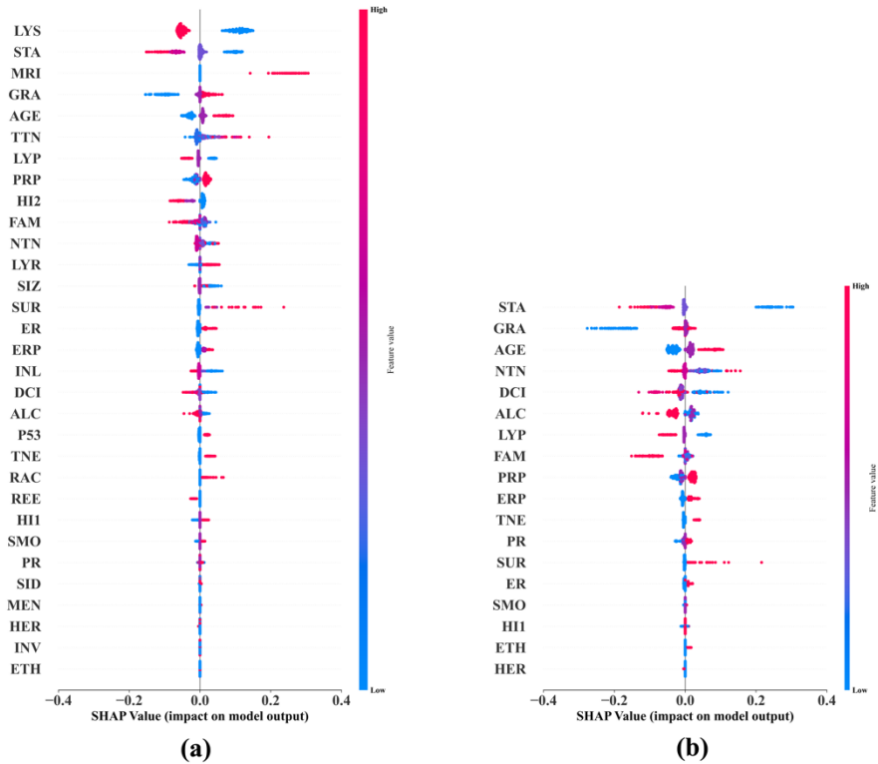

**Figure S5.** SHAP summary plots for the two best DFNN-based models concerning 10-year BCM (a) DNM-10Year; and (b) DNM_RF-10Year

*AGE: age at diagnosis of the disease; ALC: alcohol usage; DCI: type of ductal carcinoma in situ; ER: estrogen receptor expression; ERP: percent of cell stain pos for ER receptors; ETH: ethnicity; FAM: family history of cancer; GRA: grade of disease; HER: HER2 expression; HI1: tumor histology; HI2: tumor histology subtypes; INL: where invasive tumor is located; INV: whether tumor is invasive; LYP: number of positive lymph nodes; LYR: number of lymph nodes removed; LYS: patient had any positive lymph nodes; MEN: inferred menopausal status; MRI: MRIs within 60 days of surgery; NTN: number of nearby cancerous lymph nodes; PR: progesterone receptor expression; PRP: percent of cell stain pos for PR receptors; P53: whether P53 is mutated; RAC: race; REE: removal of an additional margin of tissue; SID: side of tumor; SIZ: size of tumor in mm; SMO: smoking; STA: composite of size and # positive nodes; SUR: whether residual tumor; TNE: triple negative status in terms of patient being ER, PR, and HER2 negative; TTN: prime tumor stage in TNM system.*

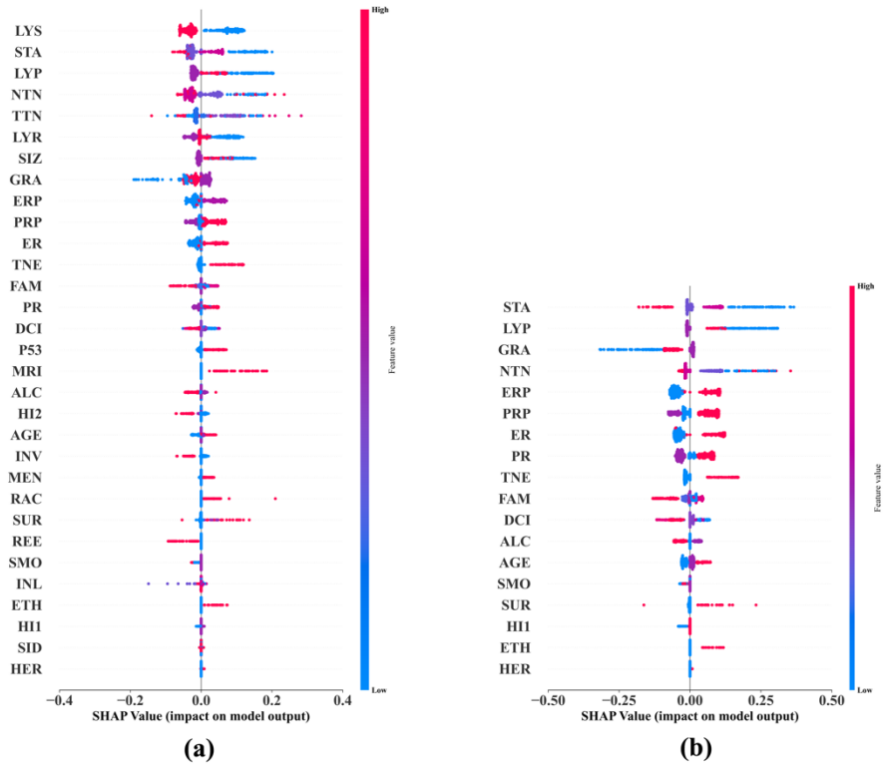

**Figure S6.** SHAP summary plots for the two best *Naïve Bayes* models concerning 10-year BCM
(a) NB-10Year; and (b) NB_RF-10Year

*AGE: age at diagnosis of the disease; ALC: alcohol usage; DCI: type of ductal carcinoma in situ; ER: estrogen receptor expression; ERP: percent of cell stain pos for ER receptors; ETH: ethnicity; FAM: family history of cancer; GRA: grade of disease; HER: HER2 expression; HI1: tumor histology; HI2: tumor histology subtypes; INL: where invasive tumor is located; INV: whether tumor is invasive; LYP: number of positive lymph nodes; LYR: number of lymph nodes removed; LYS: patient had any positive lymph nodes; MEN: inferred menopausal status; MRI: MRIs within 60 days of surgery; NTN: number of nearby cancerous lymph nodes; PR: progesterone receptor expression; PRP: percent of cell stain pos for PR receptors; P53: whether P53 is mutated; RAC: race; REE: removal of an additional margin of tissue; SID: side of tumor; SIZ: size of tumor in mm; SMO: smoking; STA: composite of size and # positive nodes; SUR: whether residual tumor; TNE: triple negative status in terms of patient being ER, PR, and HER2 negative; TTN: prime tumor stage in TNM system.*

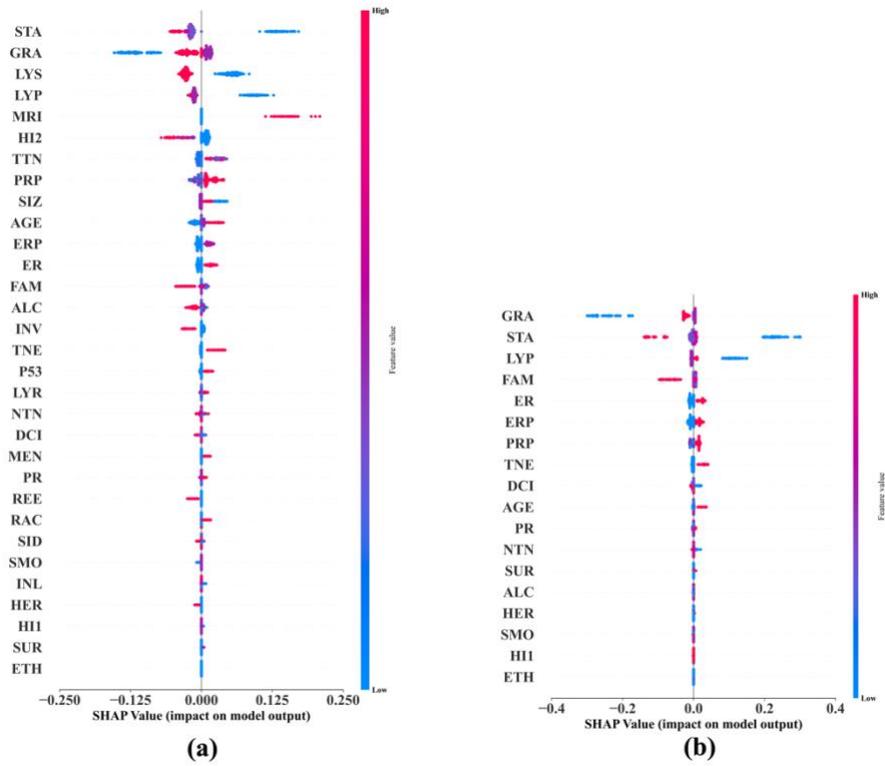

**Figure S7.** SHAP summary plots for the two best *Random Forests* models concerning 10-year BCM
(a) RaF-10Year; and (b) RaF_RF-10Year

*AGE: age at diagnosis of the disease; ALC: alcohol usage; DCI: type of ductal carcinoma in situ; ER: estrogen receptor expression; ERP: percent of cell stain pos for ER receptors; ETH: ethnicity; FAM: family history of cancer; GRA: grade of disease; HER: HER2 expression; HI1: tumor histology; HI2: tumor histology subtypes; INL: where invasive tumor is located; INV: whether tumor is invasive; LYP: number of positive lymph nodes; LYR: number of lymph nodes removed; LYS: patient had any positive lymph nodes; MEN: inferred menopausal status; MRI: MRIs within 60 days of surgery; NTN: number of nearby cancerous lymph nodes; PR: progesterone receptor expression; PRP: percent of cell stain pos for PR receptors; P53: whether P53 is mutated; RAC: race; REE: removal of an additional margin of tissue; SID: side of tumor; SIZ: size of tumor in mm; SMO: smoking; STA: composite of size and # positive nodes; SUR: whether residual tumor; TNE: triple negative status in terms of patient being ER, PR, and HER2 negative; TTN: prime tumor stage in TNM system.*

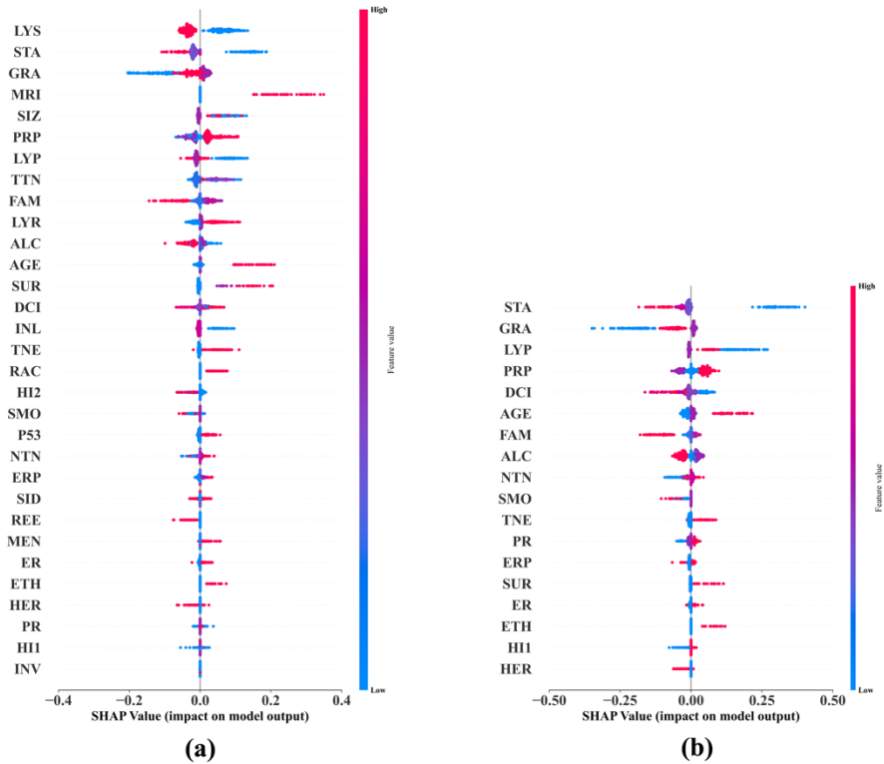

**Figure S8.** SHAP summary plots for the two best XGBoost models concerning 10-year BCM
(a) XGB-10Year; and (b) XGB_RF-10Year

*AGE: age at diagnosis of the disease; ALC: alcohol usage; DCI: type of ductal carcinoma in situ; ER: estrogen receptor expression; ERP: percent of cell stain pos for ER receptors; ETH: ethnicity; FAM: family history of cancer; GRA: grade of disease; HER: HER2 expression; HI1: tumor histology; HI2: tumor histology subtypes; INL: where invasive tumor is located; INV: whether tumor is invasive; LYP: number of positive lymph nodes; LYR: number of lymph nodes removed; LYS: patient had any positive lymph nodes; MEN: inferred menopausal status; MRI: MRIs within 60 days of surgery; NTN: number of nearby cancerous lymph nodes; PR: progesterone receptor expression; PRP: percent of cell stain pos for PR receptors; P53: whether P53 is mutated; RAC: race; REE: removal of an additional margin of tissue; SID: side of tumor; SIZ: size of tumor in mm; SMO: smoking; STA: composite of size and # positive nodes; SUR: whether residual tumor; TNE: triple negative status in terms of patient being ER, PR, and HER2 negative; TTN: prime tumor stage in TNM system.*

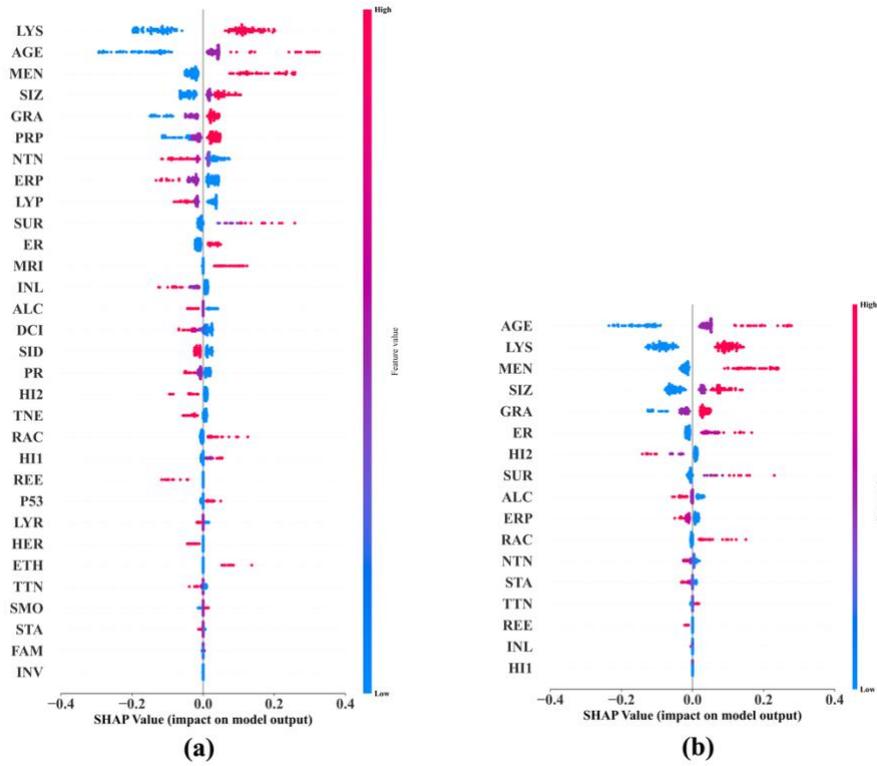

**Figure S9.** SHAP summary plots for the two best LASSO models concerning 15-year BCM
(a) LASSO-15Year; and (b) LASSO_RF-15Year

*AGE: age at diagnosis of the disease; ALC: alcohol usage; DCI: type of ductal carcinoma in situ; ER: estrogen receptor expression; ERP: percent of cell stain pos for ER receptors; ETH: ethnicity; FAM: family history of cancer; GRA: grade of disease; HER: HER2 expression; HI1: tumor histology; HI2: tumor histology subtypes; INL: where invasive tumor is located; INV: whether tumor is invasive; LYP: number of positive lymph nodes; LYR: number of lymph nodes removed; LYS: patient had any positive lymph nodes; MEN: inferred menopausal status; MRI: MRIs within 60 days of surgery; NTN: number of nearby cancerous lymph nodes; PR: progesterone receptor expression; PRP: percent of cell stain pos for PR receptors; P53: whether P53 is mutated; RAC: race; REE: removal of an additional margin of tissue; SID: side of tumor; SIZ: size of tumor in mm; SMO: smoking; STA: composite of size and # positive nodes; SUR: whether residual tumor; TNE: triple negative status in terms of patient being ER, PR, and HER2 negative; TTN: prime tumor stage in TNM system.*

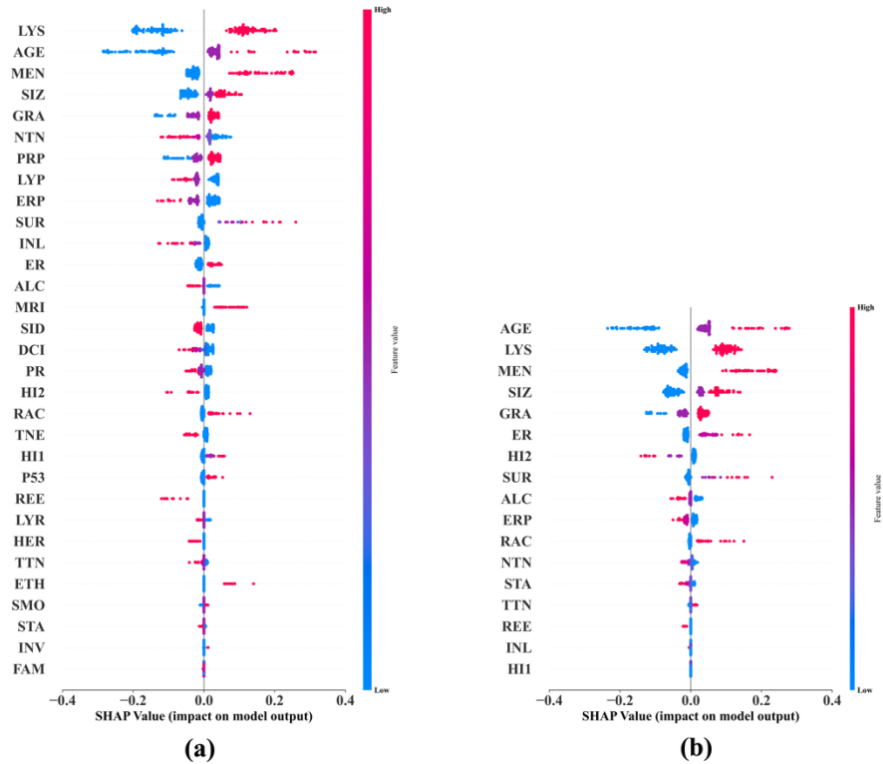

**Figure S10.** SHAP summary plots for the two best *Logistic Regression* models concerning 15-year BCM (a) LR-15Year; and (b) LR_RF-15Year

*AGE: age at diagnosis of the disease; ALC: alcohol usage; DCI: type of ductal carcinoma in situ; ER: estrogen receptor expression; ERP: percent of cell stain pos for ER receptors; ETH: ethnicity; FAM: family history of cancer; GRA: grade of disease; HER: HER2 expression; HI1: tumor histology; HI2: tumor histology subtypes; INL: where invasive tumor is located; INV: whether tumor is invasive; LYP: number of positive lymph nodes; LYR: number of lymph nodes removed; LYS: patient had any positive lymph nodes; MEN: inferred menopausal status; MRI: MRIs within 60 days of surgery; NTN: number of nearby cancerous lymph nodes; PR: progesterone receptor expression; PRP: percent of cell stain pos for PR receptors; P53: whether P53 is mutated; RAC: race; REE: removal of an additional margin of tissue; SID: side of tumor; SIZ: size of tumor in mm; SMO: smoking; STA: composite of size and # positive nodes; SUR: whether residual tumor; TNE: triple negative status in terms of patient being ER, PR, and HER2 negative; TTN: prime tumor stage in TNM system.*

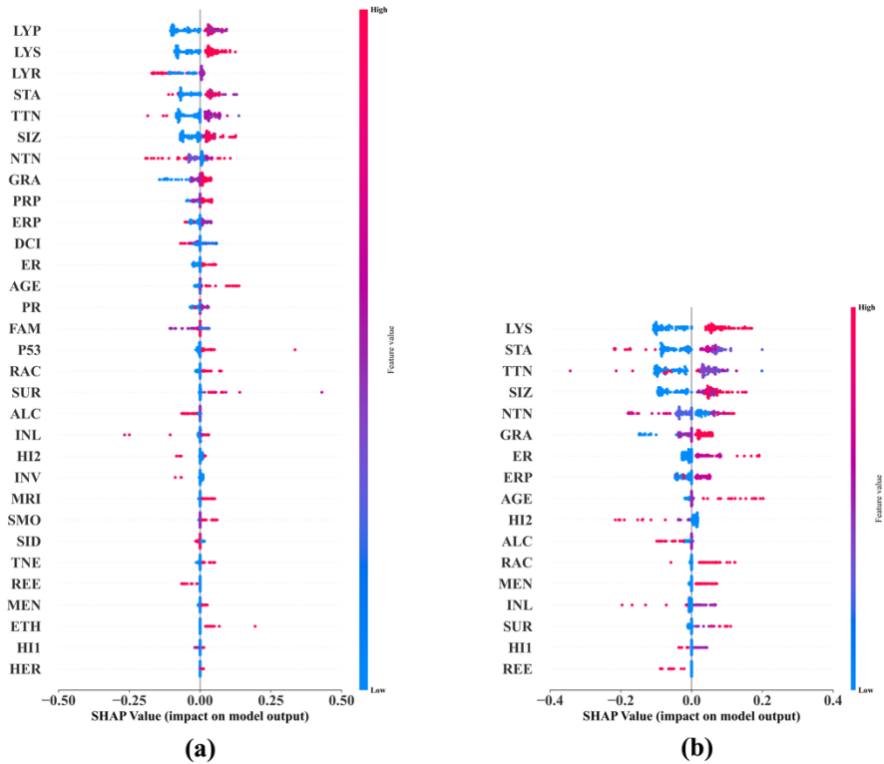

**Figure S11.** SHAP summary plots for the two best SVC models concerning 15-year datasets
(a) SVC-15Year; and (b) SVC_RF-15Year

*AGE: age at diagnosis of the disease; ALC: alcohol usage; DCI: type of ductal carcinoma in situ; ER: estrogen receptor expression; ERP: percent of cell stain pos for ER receptors; ETH: ethnicity; FAM: family history of cancer; GRA: grade of disease; HER: HER2 expression; HI1: tumor histology; HI2: tumor histology subtypes; INL: where invasive tumor is located; INV: whether tumor is invasive; LYP: number of positive lymph nodes; LYR: number of lymph nodes removed; LYS: patient had any positive lymph nodes; MEN: inferred menopausal status; MRI: MRIs within 60 days of surgery; NTN: number of nearby cancerous lymph nodes; PR: progesterone receptor expression; PRP: percent of cell stain pos for PR receptors; P53: whether P53 is mutated; RAC: race; REE: removal of an additional margin of tissue; SID: side of tumor; SIZ: size of tumor in mm; SMO: smoking; STA: composite of size and # positive nodes; SUR: whether residual tumor; TNE: triple negative status in terms of patient being ER, PR, and HER2 negative; TTN: prime tumor stage in TNM system.*